\documentclass[preprint]{elsarticle}

\usepackage{lmodern}         % Improved Computer Modern font
\journal{---}

\usepackage[utf8]{inputenc} % allow utf-8 input
\usepackage[T1]{fontenc}    % use 8-bit T1 fonts
\usepackage[hidelinks]{hyperref}    % hyperlinks
\usepackage{url}            % simple URL typesetting
\usepackage{booktabs}       % professional-quality tables
\usepackage{amsfonts}       % blackboard math symbols
\usepackage{nicefrac}       % compact symbols for 1/2, etc.
\usepackage{microtype}      % microtypography
\usepackage{xcolor}         % colors

\usepackage{adjustbox}

\usepackage{amsmath}
\usepackage{amssymb}
\usepackage{mathtools}
\usepackage{amsthm}

\usepackage[capitalize,noabbrev]{cleveref}

\usepackage{microtype}
\usepackage{graphicx}
\usepackage{subfigure}
\usepackage{booktabs} % for professional tables
\usepackage{amsfonts}
\usepackage{amssymb}
\usepackage{bm}
\usepackage{amsmath}
\usepackage{amsthm}
\usepackage{stmaryrd}
\usepackage{color}
\usepackage{xcolor}
\usepackage[subnum]{cases}
\usepackage{rotating}
\usepackage{enumitem}
\usepackage{accents}
\usepackage[title]{appendix}
\usepackage{mathtools}
\usepackage{caption}
\usepackage{url}
\usepackage{framed}
\usepackage{lscape}
\usepackage{multirow}
\usepackage{latexsym}
\usepackage{mathrsfs}
\usepackage{array}
\usepackage{makecell}
\usepackage{float}
\usepackage{tabularray}% for long table
\usepackage{tablefootnote}
\usepackage{tikz}
\usepackage[all]{xy}
\usepackage{tikz-cd}
\usepackage[usestackEOL]{stackengine}
\usepackage{pdflscape}
\usepackage{siunitx}
\usepackage{wrapfig}
\usepackage{comment}

\usepackage{hyperref}

\theoremstyle{plain}
\newtheorem{theorem}{Theorem}[section]
\newtheorem{proposition}[theorem]{Proposition}

\newtheorem{corollary}[theorem]{Corollary}
\newtheorem{remark}[theorem]{Remark}

\theoremstyle{definition}
\newtheorem{definition}[theorem]{Definition}

\newcommand{\Cl}{\operatorname{Cl}}
\newcommand{\Ortho}{\operatorname{O}}

\renewcommand{\le}{\leqslant}
\renewcommand{\ge}{\geqslant}
\renewcommand{\leq}{\leqslant}
\renewcommand{\geq}{\geqslant}
\newcommand{\Rbb}{\mathbb{R}}

\newcommand{\GL}{\operatorname{GL}} % nhóm tuyến tính tổng quát

\newcommand{\OO}{\operatorname{O}}
\newcommand{\T}{\operatorname{T}}

\newcommand{\id}{\operatorname{id}}

\newcommand{\E}{\operatorname{E}}

\newcommand{\Aut}{\operatorname{Aut}}

\newcommand{\nf}{\operatorname{nf}}

\begin{document}
\begin{frontmatter}

% Title
\title{A Clifford Algebraic Approach to E(n)-Equivariant High-order Graph Neural Networks}

% Authors and affiliations
\author{Viet-Hoang Tran\fnref{NUS}\corref{contrib}%
}
\author{Thieu N. Vo\fnref{NUS}\corref{contrib}%
}
\author{Tho Tran Huu\fnref{NUS}%
}
\author{Tan Minh Nguyen\fnref{NUS}%
}
\fntext[NUS]{National University of Singapore%
}
\cortext[contrib]{Equal contributions.}

\begin{abstract}
Designing neural network architectures that can handle data symmetry is crucial.  This is especially important for geometric graphs whose properties are equivariance under Euclidean transformations. Current equivariant graph neural networks (EGNNs), particularly those using message passing, have a limitation in expressive power. Recent high-order graph neural networks can overcome this limitation, yet they lack equivariance properties, representing a notable drawback in certain applications in chemistry and physical sciences. In this paper, we introduce the Clifford Group Equivariant Graph Neural Networks (CG-EGNNs), a novel EGNN that enhances high-order message passing by integrating high-order local structures in the context of Clifford algebras. As a key benefit of using Clifford algebras, CG-EGNN can learn functions that capture equivariance from positional features. By adopting the high-order message passing mechanism, CG-EGNN gains richer information from neighbors, thus improving model performance. Furthermore, we establish the universality property of the $k$-hop message passing framework, showcasing greater expressive power of CG-EGNNs with additional $k$-hop message passing mechanism. We empirically validate that CG-EGNNs outperform previous methods on various benchmarks including n-body, CMU motion capture, and MD17, highlighting their effectiveness in geometric deep learning.
\end{abstract}
\end{frontmatter}

% \maketitle

\section{Introduction}
\label{sec:introduction}

%\tvh{Outline of Introduction:
%- Talk a lots .... (no equation!)
%- Contribution.
%- Notation.
%}

Developing neural network architectures capable of accommodating the symmetry constraints inherent in data and transformations is essential in geometric deep learning and remains a highly active research domain~\citep{pozdnyakov2022incompleteness, hodapp2023equivariant, batatia2024general}.
% Designing neural network architectures that can handle the symmetry constraints of data and transformations is crucial in geometric deep learning and remains a highly active research domain. 
This is especially important for graph-based applications, such as those in chemistry~\cite{klicpera2021gemnet, batzner20223, musaelian2023learning} and physical sciences~\cite{batatia2022mace, batatia2022design, bochkarev2022multilayer, nigam2022unified}, where nodes in the graph represent points in Euclidean space and the properties of the graph stay equivariant under Euclidean transformations. 
Convolutional Neural Networks (CNNs), which exhibit translation equivariance, and Graph Neural Networks (GNNs), which demonstrate permutation equivariance, %[~\tanr{GNNs is not always permutation invariant. Should it be equivariant GNNs}], 
are notable examples of the significant success and effectiveness of integrating symmetry-aware structure into neural network architectures \citep{bruna2013spectral, defferrard2016convolutional, kipf2016semi}.

% Numerous approaches for integrating equivariant properties are available in the literature, drawing from a diverse array of mathematical formulations\tanm{~\cite{}}. 
% These approaches can be categorized into three distinct families: scalarization, regular group representations, and irreducible representations.
% Scalarization manipulates scalar features or vectors through multiplication\tanm{~\cite{}}. 
% Regular representations are effective for finite groups, relying on integrals over the group\tanm{~\cite{}}. 
% Irreducible representations, usually used for $SO(3)$ or $O(3)$ equivariance, employ Wigner-$D$ matrices and Clebsch-Gordan coefficients within a spherical harmonics basis\tanm{~\cite{}}. 
% However, all of these methods have certain limitations, such as the inability to capture all directional information, \tanm{the challenge of establishing alternative bases, and high computational complexity}\tanm{~\cite{}}.
% Recently, emerging approaches leverage noncommutative algebras like Hamiltonian quaternions and geometric and Clifford algebras to construct a new kind of equivariant networks. 
% These networks operate directly on vector bases, bypassing alternative basis representations, thereby simplifying the action of orthogonal transformations through algebraic multiplication\tanm{~\cite{}}. 
% This approach not only facilitates computational geometry without matrices or tensors but also holds promise for advancing equivariant neural network design.

The $\E(n)$-equivariant Graph Neural Network (EGNN) model, as proposed by \cite{satorras2021n}, is specifically designed for graph data and can be considered as a scalarization approach. %[\tanr{What is scalarization? This term is not common in ML. Please elaborate here}]. 
EGNN finds broader applications in drug design, molecular modeling, and 3D point cloud primarily due to its efficiency and straightforward model design \cite{brandstetter2021geometric, kohler2019equivariant, schutt2017schnet}. 
In principle, EGNN adopts the message passing framework from GNNs and additionally incorporates the information of the distances of nodes into the message update in such a way that the equivariant property is achieved \cite{trang20243, eijkelboom2023n}.
%Building upon this foundation, the research conducted by \citep{brandstetter2021geometric} extends EGNN's capabilities by incorporating insights from spherical-harmonics-type architectures. 
%Their creation, the Steerable E(n)-Equivariant Graph Neural Network (SEGNN), demonstrates superior performance across certain benchmarks. 
%However, it shares similar conceptual limitations with EGNN while also introducing increasing computational complexity.

%[\tanr{We need much more citations in the first three paragraph in the intro. PLEASE CITE}]
Despite their effectiveness, EGNNs have %[\tanr{We should use EGNN instead of EGNNs but EGNNs ok for now}] 
limited expressive power due to its inability to distinguish individual nodes. 
To tackle this challenge, researchers have introduced several high-order GNNs. These models leverage an encoding of $k$-tuples of nodes, subgraphs, or hyper-graph and then apply either message passing techniques~\citep{morris2019weisfeiler} or equivariant tensor operations~\citep{maron2018invariant}. 
High-order GNNs hold the promise of incorporating richer information from local subgraphs into the message passing process, thereby achieving superior results compared to traditional GNNs. 
Nevertheless, it is worth noting that despite their enhanced capabilities, these high-order GNNs lack equivariance with respect to %[\tanr{are not equivariant to}] 
Euclidean transformations. This absence of equivariance %[\tanr{such an equivariant property}]
poses a significant drawback in specific applications within the fields of drug design, chemistry and physical sciences. %[\tanr{many relevant applications, such as drug design, molecular modeling, and 3D point cloud}] [Thieu: I moved to the second pararaph].

{\bf Contributions.} In this paper, we introduce a novel class of equivariant graph neural network, named the \emph{Clifford Group Equivariant Graph Neural Networks} (CG-EGNNs). 
Similar to EGNNs, our approach is based on the message passing mechanism. 
However, in contrast to EGNNs, \emph{CG-EGNNs enhance the message passing process by integrating high-order local structures around graph nodes within the framework of Clifford algebras}. 
In addition, we prove that the $k$-hop message passing mechanism satisfies the universality property for geometric graphs. 
Therefore, CG-EGNNs with the additional $k$-hop message mechanism have an ability of gathering richer information from neighboring nodes during each feature update, thereby enhancing the expressive power of the networks while preserving equivariant properties.
%This modification enables CG-EGNNs to gather richer information from neighboring nodes during each feature update, thereby enhancing the expressive power of the networks while preserving equivariant properties. 
Furthermore, in our formulation, positional features are updated only once at the final layer, eliminating the need for updating positional features at every layer and consequently reducing computational complexity. Our contribution is three-fold.
\begin{enumerate}[leftmargin=20pt]
    \item We introduce a novel class of equivariant high-order graph neural networks, namely CG-EGNNs, which enables equivariance properties of high-order message passing process by integrating high-order local structures around graph nodes within the framework of Clifford algebras.
    
    \item We theoretically prove that CG-EGNN is $\E(n)$-equivariant and capable of learning functions that capture equivariance from positional features at the same time.
     
     \item By adapting the high-order message passing mechanism, CG-EGNN can gain richer information from neighbors to each node feature updates,   resulting in performance improvements. 
     In addition, we establish the universality property of the $k$-hop message passing framework, indicating that CG-EGNN with $k$-hop message passing mechanism possesses greater expressive power.
\end{enumerate}
We demonstrate the superiority of our model over previous approaches
through significant empirical improvements on three benchmarks: n-body system, CMU motion capture dataset~\citep{cmu2003motion}, and MD17 molecular dataset~\citep{chmiela2017machine}.

\noindent \textbf{Organization.}
We structure this paper as follows: After summarizing related work in Section~\ref{sec:related_work}, we recall necessary definitions and constructions from Equivariant Graph Neural Networks and Clifford Group Equivariant Neural Networks in Section~\ref{sec:preliminaries}.
In Section~\ref{sec:methodology}, we present the detailed construction of CG-EGNNs and discuss their equivariance the importance of Clifford Algebra in CG-EGNNs.
In Section~\ref{sec:k-hop}, we propose an addition component to CG-EGNNs using $k$-hop message passing framework.
We theoretically prove the universality of the $k$-hop message passing framework for geometric GNNs, thus enhancing the expressive power of the obtained CG-EGNNs.
In Section~\ref{sec:experiments}, we conduct experiments to justify the advantages of CG-EGNNs over previous methods in the literature.
The paper ends with concluding remarks. Experimental details are provided in the Appendix.

%\tanr{We structure this paper as follows: In Section~\ref{sec:mom_smoe}, we establish the connection between SMoE and gradient descent and derive our MomentumSMoE. In Section~\ref{sec:analyis}, we theoretically prove the stability advantage of MomentumSMoE over SMoE. In Section~\ref{sec:advanced_mom_smoe}, we introduce AdamSMoE and Robust MomentumSMoE. In Section~\ref{sec:experiment}, we present our experimental results to justify the advantages of our momentum-based SMoE models over the traditional SMoE and other SMoE baselines. In Section~\ref{sec:empirical_analyis}, we empirically analyze our MomentumSMoE. We discuss related works in Section~\ref{sec:related_work}. The paper ends with concluding remarks. More experimental details are provided in the Appendix.}

\section{Related Work}\label{sec:related_work}

\textbf{Equivariant neural networks.}
The equivariance property of neural networks has been achieved through various ways, with most falling into three distinct classes: scalarization methods, regular group representations, and irreducible representations \citep{han2022geometrically}. 
Scalarization methods, such as those manipulating scalar features or vectors through scalar multiplication, have been employed, yet they often struggle to capture all directional information \citep{coors2018spherenet, kohler2020equivariant, deng2021vector, tholke2022torchmd}. 
Regular representation methods construct equivariant maps via integrals over the group under consideration \citep{cohen2016group, kondor2018generalization, finzi2020generalizing, bekkers2019b}. 
However, for infinite or continuous groups, the intractability of such integrals necessitates approximations that can compromise equivariance.

Irreducible representation methods, specifically designed for neural networks equivariant to SO(3) or O(3), utilize Wigner-D matrices and Clebsch-Gordan coefficients within a steerable spherical harmonics basis \citep{thomas2018tensor, fuchs2020se}. 
While promising, these methods face challenges in establishing an alternative base and computing Clebsch-Gordan coefficients, which are nontrivial \citep{alex2011numerical}. 
Recent approaches that leverage noncommutative algebras like Hamiltonian quaternions, geometric, and Clifford algebras offer a fresh perspective \citep{shen20203d, shen2024rotation, zhao2020quaternion, brehmer2023geometric, ruhe2023clifford}. 
Similar to scalarization methods, these approaches operate directly on vector bases, simplifying orthogonal transformations through algebraic multiplication.
Thus, they have the potential to advance equivariant neural network design.

\textbf{Equivariant graph neural networks.} 
Among the most commonly used GNN architectures are Message Passing Graph Neural Networks, which iteratively propagate messages to compute graph representations \citep{kipf2016semi, gilmer2017neural, xu2018powerful}. 
Leveraging this framework, several rotational equivariant neural networks tailored for geometric graphs have been developed, exemplified by works such as those by \cite{gasteiger2020directional, schutt2021equivariant, brandstetter2021geometric}. 
Additionally, approaches similar to equivariant multilayer perceptrons have been proposed for specialized tasks involving molecules and protein structures, showcased in studies by \cite{schutt2018schnet, gasteiger2021gemnet, jing2020learning, huang2022equivariant, anderson2019cormorant, batzner20223}.

\noindent \textbf{High-order message GNNs.}
A few high-order Graph Neural Networks (GNNs) have been proposed to enhance the expressive capabilities of traditional GNNs. 
\cite{morris2019weisfeiler} introduce a message passing mechanism tailored for k-tuples of nodes. 
In their initialization step, each k-tuple is labeled based on the isomorphism types of their induced subgraphs, ensuring distinct labels for differing subgraph structures \citep{maron2019provably}. 
Another category of high-order networks employs linear equivariant operations, interleaved with coordinate-wise nonlinearities, operating on order-k tensors comprising adjacency matrices and node attributes \citep{maron2018invariant, maron2019universality, maron2019provably}. 
These GNNs exhibit expressive power comparable to k-GNNs and are adept at counting substructures within graphs. 
However, none of these models are explicitly designed to maintain equivariance to transformations in Euclidean spaces.

\textbf{Clifford Algebra.} 
A Clifford algebra is an algebra generated by a quadratic vector space modulo some relations about the square of a vector. 
This is a generalization of real numbers, complex numbers, and a number of hypercomplex number systems such as quaternions, octonions, exterior algebra, etc. \citep{hamilton1866elements, grassmann1862ausdehnungslehre}.
It is often called geometric algebra when the base quadratic space is over the real numbers \citep{ablamowicz2004lectures}. 
Clifford algebra provides a powerful language for science and engineering that clearly describes the geometric symmetries of physical space and spacetime \citep{bayro2006conformal, hildenbrand2008inverse, wareham2004applications, breuils2022new, bhatti2021advanced, hitzer2010interactive}. 
It simplifies the action of orthogonal transformations on quadratic space through algebraic multiplication, and more generally, enables computational geometry without involving matrices or tensors \citep{melnyk2021embed, spellings2021geometric, brehmer2023geometric, liu2024clifford}. 
%{\color{red} Need citation}
%https://davidruhe.github.io/2023/06/14/clifford-group.html

\section{Preliminaries} 
\label{sec:preliminaries}

{\bf Equivariant Neural Networks.} Given two sets $X, Y$ and a group $G$ acting on them, a function $\phi \colon X \to Y$ is called $G$-equivariant if $\phi(g \cdot x) = g \cdot \phi(x)$ for all $x \in X$ and $g \in G$. 
If $G$ acts trivially on $Y$, then we say $\phi$ is $G$-invariant. 

{\bf Message Passing Mechanism.} Given a graph $\mathcal{G}=(\mathcal{V},\mathcal{E})$ with $M$ nodes $i \in \mathcal{V}$ and edges $e_{i,j} \in \mathcal{E}$. Each node $i \in \mathcal{V}$ is associated with a node feature embedding $\mathbf{h}_i \in \mathbb{R}^{\nf}$. Message Passing Mechanism refers to sharing information between nodes in a graph along the edges. Following the notation in \cite{pmlr-v70-gilmer17a}, the message passing layer can be presented as follows:
\begin{align*}
    \mathbf{m}_{i,j} &= \phi_m(\mathbf{h}_i^{l},\mathbf{h}_j^{l},e_{i,j}),\quad \mathbf{h}_i^{l+1} = \phi_h \left(\mathbf{h}_i^{l},\sum\limits_{j \in \mathcal{N}(i)} \mathbf{m}_{i,j} \right),
\end{align*}
where $\mathbf{h}^l_i \in \mathbb{R}^{\nf}$ is the node feature embedding of node $i$ at layer $l$. $a_{ij}$ is the edge attribute. Here, $\phi_m, \phi_h$ are learnable neural networks. 

{\bf Equivariant Graph Neural Networks.} In the original setting of EGNN \citep{satorras2021n}, each nodes $i \in \mathcal{V}$ is additionally associated with $\mathbf{x}_i \in \mathbb{R}^n$ as an $n$-dimensional coordinate embedding.  
The main component of EGNN is the Equivariant Graph Convolution Layer (EGCL) which takes $(\mathbf{x}^{l},\mathbf{h}^{l})$ as input and outputs $(\mathbf{x}^{l+1},\mathbf{h}^{l+1})$, as follows:
\begin{align}
    \mathbf{m}_{i,j} &= \phi_m\left(\mathbf{h}_i^{l},\mathbf{h}_j^{l},\|\mathbf{x}_i^{l}-\mathbf{x}_j^{l}\|_2^2,e_{i,j}\right), \label{L2inEGNN}\\
	\mathbf{x}_i^{l+1} = \mathbf{x}_i^{l} + \frac{1}{M-1} \sum\limits_{j \neq i} &(\mathbf{x}_i^{l}-\mathbf{x}_j^{l}) \phi_x(\mathbf{m}_{i,j}), \quad \mathbf{h}_i^{l+1} = \phi_h \left(\mathbf{h}_i^{l},\sum\limits_{j \in \mathcal{N}(i)} \mathbf{m}_{i,j} \right). \nonumber
 %\label{eq:EGNN-update-x}
%\label{eq:EGNN-update-h}
\end{align}
In \cite{satorras2021n}, it has been proved that EGCL is $\E(n)$-equivariant, i.e., $Q \cdot \mathbf{x}^{l+1}+g,\mathbf{h}^{l+1} = \text{EGCL}(Q \cdot \mathbf{x}^{l}+g,\mathbf{h}^{l})$,
% \begin{equation}
% \label{eq:EGNN-equivariance}
% 	Q \cdot \mathbf{x}^{l+1}+g,\mathbf{h}^{l+1} = \text{EGCL}(Q \cdot \mathbf{x}^{l}+g,\mathbf{h}^{l}),
% \end{equation}
for all orthogonal matrix $Q \in \OO(n)$ and translation vector $g \in \mathbb{R}^n$.

{\bf Clifford Algebra and Clifford Group Equivariant Neural Networks.} Let $\mathbb{F}$ be a field with $\operatorname{char}\mathbb{F} \neq 2$. Clifford Algebra \cite{Garling_2011, ruhe2023clifford}, denoted as $\Cl(V,\mathfrak{q})$ where $(V,\mathfrak{q})$ is a quadratic space over $\mathbb{F}$, is the $\mathbb{F}$-algebra generated by generated by $V$ with relations $v^2 = \mathfrak{q}(v)$ for all $v \in V$. The authors in \cite{ruhe2023clifford} introduced a variant of Clifford Group and provided a class of neural networks, named Clifford Group equivariant neural networks (CGENN), that operate on elements of Clifford Algebra. It is worth noting that, when $(V,\mathfrak{q})$ is the vector space $\mathbb{R}^n$ with quadratic form $\mathfrak{q}$ is the square of the Euclidean norm, i.e.,  $\mathfrak{q}(\cdot) = \| \cdot \|_2^2$, the corresponding Clifford group is closely related to the orthogonal group $\Ortho(n)$, defined as
\begin{equation}
    \Ortho(n) = \{ Q \in \GL(n) ~ | ~ Q^{\top}Q = QQ^{\top} = I_n \}.
\end{equation}
%It [\tanr{What is it here?}] 
This relation between the Clifford group and $\Ortho(n)$ allows us to leverage CGENNs to build $\Ortho(n)$-equivariant or invariant neural networks.

\section{Clifford Group Equivariant Graph Neural Networks}
\label{sec:methodology}

In this section, we present our CG-EGNNs by extending EGNNs in the context of Clifford algebras.
We then discuss further improvements using high-order inputs.

\subsection{Clifford Group Equivariant Graph Neural Networks (CG-EGNNs)}
\begin{figure*}
  \begin{center}
  % \vskip -0.1in
    \includegraphics[width=0.95\textwidth]{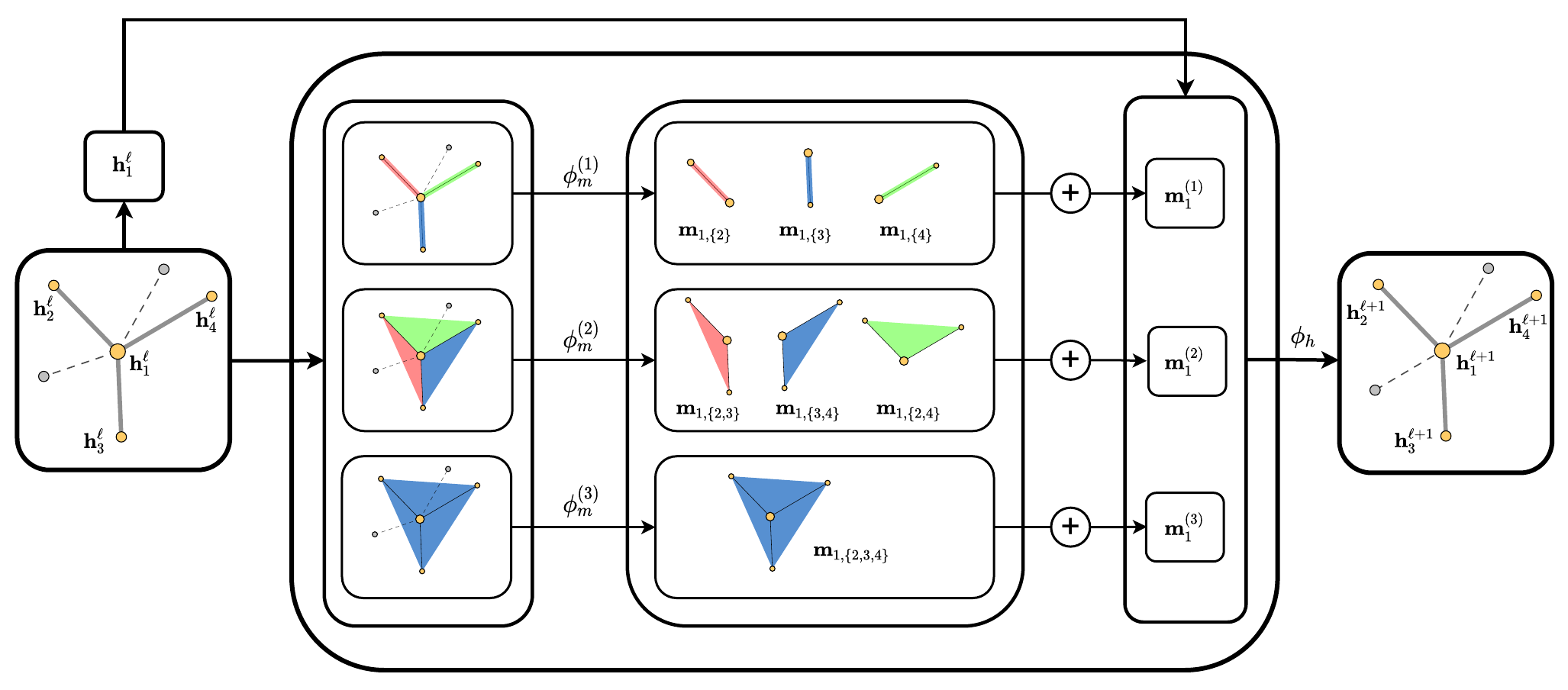}
  \end{center}
  % \vskip-0.25in
  \caption{Illustration of high-order message passing mechanism (Eqs.~\eqref{partialmessage}-\eqref{featurecompute}) in CG-EGNN. Here, the feature of node $1$ is updated by computing and aggregating messages of order $d=1,2,3$ from its neighborhood $\mathcal{N}(1) = \{2,3,4\}$.}
  \label{mainfigure}
  \vskip-0.1in
\end{figure*}
Following notations from the original setting of EGNNs, we consider a graph $\mathcal{G} = (\mathcal{V}, \mathcal{E})$ with $M$ nodes $i \in \mathcal{V}$ and edges $e_{i,j} \in \mathcal{E}$.
To present the architecture of our model, let us assume that each node $i \in \mathcal{V}$ is associated with an initial position $\mathbf{x}_i \in \mathbb{R}^n$, some vector features $\mathbf{v}_i=(\mathbf{v}_{i,1}, \ldots, \mathbf{v}_{i,r}) \in \mathbb{R}^{n \times r}$,
and some scalar features $a_i=(a_{i,1},\ldots,a_{i,s}) \in \mathbb{R}^s$.
Depending on particular experiments, vector features can be either velocity or acceleration, while scalar features can be mass, charge, temperature, and so on. 

CG-EGNNs will preserve equivariance on the set of vector features and invariance on the set of scalar features to Euclidean group $\E(n)$. It has three main components: 
	the embedding layer,
 	the Clifford graph convolution layer,
 	and the projection layer.

\textbf{Embedding Layer.} This layer transforms the initial information at each node into multivector $\mathbf{h}_i^0 \in \Cl({\mathbb{R}^n})^{\nf}$, of the Clifford algebra $\Cl(\mathbb{R}^n)$.
By using the identifications $\Cl(\mathbb{R}^n)^{(0)} \equiv \mathbb{R}$ and $\Cl(\mathbb{R}^n)^{(1)} \equiv \mathbb{R}^n$, 
we can view the position $\mathbf{x}_i$ and the vector features $\mathbf{v}_{i,j}$ as elements of order $1$, and the scalar features $a_{i,j}$  as elements of order 0 of the Clifford algebra $\text{Cl}(\mathbb{R}^n)$.
The output $\mathbf{h}_i^0$ of this layer is  determined by:
\begin{equation}
\label{embeddinglayer}
    \mathbf{h}_i^0 = \phi_{\operatorname{embed}} (\mathbf{x}_i,\mathbf{v}_{i,1},\ldots,\mathbf{v}_{i,r},a_{i,1},\ldots,a_{i,s}).
\end{equation} 
Here, $\phi_{\operatorname{embed}}$ is a learnable Clifford neural network.

\textbf{Clifford Graph Convolution Layer.} 
The Clifford graph convolution layer takes as input the multivector features $\mathbf{h}^{l} = \{\mathbf{h}_i^{l}\}_{i \in \mathcal{V}}$ and edge information $\mathcal{E}=\{e_{i,j}\}$. Its outputs are the multivector features $\mathbf{h}^{l+1} \in \Cl(\mathbb{R}^n)^{\nf}$.
In contrast to the original EGNNs, we incorporate not only messages from neighboring nodes but also messages from higher-dimensional neighbor structures into the update of multivector features $\mathbf{h}^{l+1}$.
In particular, fix a positive integer $D$ representing the highest-order of the neighbor structures that we want to incorporate.
For each $d=1,\ldots,D$, and each subset $A \subseteq \mathcal{N}(i)$ with $d$ elements, we determine the message from $A$ contributing to the node $i$ as follows:
\begin{align}
\label{partialmessage}
    \mathbf{m}_{i,A} = \phi_m^{(d)} \left( \mathbf{h}_i^{l},\sum_{j \in A} \mathbf{h}_j^{l} \right).
\end{align}
We can optionally add information about edge attributes to increase expressivity by concatenating them to the input of the function in a suitable way. Then, we incorporate these messages to the multivector feature $\mathbf{h}_i^{l+1}$ as follows:
\begin{align}
    \mathbf{m}_i^{(d)} & = \sum_{A \subseteq \mathcal{N}(i),\, |A|=d} \mathbf{m}_{i,A},\quad d=1,\ldots,D, \label{computehigherordermessage}\\
        \mathbf{h}_i^{l+1} &= \phi_h 
(\mathbf{h}_i^{l},\mathbf{m}_{i}^{(1)},\ldots,\mathbf{m}_i^{(D)}). \label{featurecompute}
\end{align}
Here, $\phi_m^{(d)}$'s and $\phi_h$ are learnable Clifford neural networks. A visualization of a Clifford Graph Convolution Layer is given in Figure~\ref{mainfigure}.

\textbf{Projection Layer.} The embedding layer let us embed input into a multivector $\mathbf{h}^0_i \in \Cl(\mathbb{R}^n)^{\nf}$. We now need to extract information from the obtained multivector $\mathbf{h}_i^L \in \Cl(\mathbb{R}^n)^{\nf}$ to get the output of the model. This is based on the objects to 
 be predicted. If we want to predict the vector features, for example, the final position $\mathbf{x}^L_i \in \Cl^{(1)}(\mathbb{R}^n)  \equiv \mathbb{R}^n$, we compute it as follows:
\begin{align} \label{updateposition}
    \mathbf{x}_i^L = \mathbf{x}_0 + \phi_x \left (\mathbf{h}_i^L \right)^{(1)},
\end{align}
where $\cdot ^{(1)}$ is the projection of $\Cl(\mathbb{R}^n)$ onto subspace $\Cl^{(1)}(\mathbb{R}^n)$. Similarly, if we want to predict some scalar information of vertices, let say $\mathbf{output}_i^L \in \Cl^{(0)}(\mathbb{R}^n)  \equiv \mathbb{R}$, we use the projection $\cdot ^{(0)}$ of $\Cl(\mathbb{R}^n)$ onto subspace $\Cl^{(0)}(\mathbb{R}^n)$ and compute the result as follows:
\begin{align}
    \mathbf{output}_i^L = \phi_{\operatorname{output}}\left(\mathbf{h}_i^L \right)^{(0)}.
\end{align}

Here, $\phi_x$ and $\phi_{\operatorname{output}}$ are learnable linear Clifford layers. Note that we specifically use different projections of the Clifford algebra base on we want the model to be invariant or equivariant.

Overall, the CG-EGNN is a composition of the embedding layer, $L$ Clifford graph convolution layers, and the projection layer.

\begin{remark}
    To archive $\E(n)-$equivariance, at the embedding layer, we input the mean-subtracted positions. This maintains translation invariance. In some tasks that we want to predict position information, the residual connection in Eq.~\eqref{updateposition} maintains translation equivariance.
\end{remark}

\subsection{E(n)-equivariant Property of CG-EGNNs}
We consider our CG-EGNN consists of the embedding layer in Eq.~\eqref{embeddinglayer}, $L$ Clifford graph convolution layers in Eq.~\eqref{partialmessage},~\eqref{computehigherordermessage}, and~\eqref{featurecompute}, and finally the projection layer in Eq.~\eqref{updateposition}. We also center the input by taking the mean-subtracted position of $\mathbf{x}_i$ in Eq.~\eqref{embeddinglayer}. We provide formal proofs for the below statements in Appendix~\ref{appendix:equivariantproof}.
% Other options for CG-EGNN construction share the same properties and similar proofs.

\begin{proposition}
\label{theorem:equivariance}
The following maps are $\Ortho(n)$-equivariant:
\begin{enumerate}
    \item The map in Eq.~\eqref{embeddinglayer}:
        \begin{align}
            \mathbf{Embedding} \colon \Cl(\mathbb{R}^n)^{1+r+s} & \longrightarrow \Cl(\mathbb{R}^n)^{\nf} \notag \\
            (\mathbf{x}_i, \mathbf{v}_{i}, a_i) \quad & \longmapsto \mathbf{h}_i^0.
        \end{align}
    \item The map in Eq.~\eqref{partialmessage}, Eq.~\eqref{computehigherordermessage}, Eq.~\eqref{featurecompute}:
\begin{align}
    \mathbf{Convolution} \colon \Cl(\mathbb{R}^n)^{\nf} & \longrightarrow \Cl(\mathbb{R}^n)^{\nf} \notag \\
    \mathbf{h}_i^{l} \quad & \longmapsto \mathbf{h}_i^{l+1}.
\end{align}
    \item The map in Eq.~\eqref{updateposition}:
\begin{align}
    \mathbf{Projection} \colon \Cl(\mathbb{R}^n)^{\nf} & \longrightarrow \Cl(\mathbb{R}^n) \notag \\
    \mathbf{h}_i^L \quad & \longmapsto \phi_x(\mathbf{h}_i^L)^{(1)}.
\end{align}
\end{enumerate}
\end{proposition}

\begin{remark}
Note that $\mathbf{h}_i^0$ is $\Ortho(n)$-equivariant to $(\mathbf{x}_i, \mathbf{v}_{i}, a_i)$; $\mathbf{m}_{i,A}, \mathbf{m}_i^{(d)}, \mathbf{h}_i^{l+1}$ are $\Ortho(n)$-equivariant to $\mathbf{h}_i^{l}$; and $\phi_x(\mathbf{h}^L)^{(1)}$ is $\Ortho(n)$-equivariant to $\mathbf{h}_i^L$.
\end{remark}

\begin{remark}
Inductively, a composition of these above layers will also be $\Ortho(n)$-equivariant.
\end{remark}

By the semidirect product $\E(n) = \T(n) \rtimes \Ortho(n)$, where $\T(n)$ is the translational group, inputting the mean-subtracted positions and composing with residual connection in Eq.~\eqref{updateposition} makes CG-EGNN achieves translation equivariance. 
In particular, messages $\mathbf{m}_{i,A}, \mathbf{m}_i^{(d)}$ and features $\mathbf{h}_i^{l}$ are $\T(n)$-invariant, and position $\mathbf{x}_i^L$ is $\T(n)$-equivariant to initial position $\mathbf{x}_i$. Hence, our model becomes $\E(n)$-equivariant. We summarize our results by the following corollary.
\begin{corollary}
\label{theorem:enequivariant}
CG-EGNN is $\E(n)$-equivariant. Concretely, we have:
\begin{equation}
    \label{eq:equivarianceCGEGNN}
	Q \cdot \mathbf{x}^{L}+g = \operatorname{CG-EGNN}(Q \cdot \mathbf{x}+g),
\end{equation}
for all orthogonal matrix $Q \in \Ortho(n)$ and translation vector $g \in \mathbb{R}^n$. 
\end{corollary}

\subsection{Learning Functions that Capture Positional Equivariance by Clifford Algebras}
%\subsection{Comparison with Previous Works}

As one of the key benefits of using Clifford algebra, our CG-EGNNs can learn the component that captures positional equivariance and invariance from the input.
Indeed, the main component that captures equivariance in many previous equivariant graph neural network architectures \cite{thomas2018tensor, kohler2020equivariant, schutt2017schnet, satorras2021n} is the Euclidean norm $\|\mathbf{x}_i-\mathbf{x}_j\|^2_2$ given in Eq.~\eqref{L2inEGNN}, which is $\E(n)$-invariant. 
In CG-EGNNs, instead of fixing such a component from the input, we directly embed inputs into the Clifford algebra and let the CG-EGNN maintain equivariance by learning implicit components that capture both positional invariance and equivariance by itself. 
%This makes our model more flexible than previous models.
%In CG-EGNNs, instead of fixing such a component from the input, we directly embed each $\mathbf{x}_i$ then compute $\mathbf{h}^0_i$, since the architecture of the  Clifford neural network itself maintains equivariance. 
%This makes our model more flexible than previous models since ours implicitly learns a function that will capture equivariance from position information $\mathbf{x}_1, \ldots, \mathbf{x}_M$. 

The implicit components that capture positional equivariance and invariance learned by CG-EGNNs are quite flexible since the Clifford layers presented in \cite{ruhe2023clifford} provide a good class of learnable maps that is $\Ortho(n)$-equivariant and invariant.
For example, all polynomials with real coefficients are $\Ortho(n)$-equivariant.
%This comes from the Clifford layers presented in \cite{ruhe2023clifford} provide a good class of learnable maps that is $\Ortho(n)$-equivariant; for example, polynomials with real coefficients.
The Euclidean norm, which is the fixed component used in the existing models, is a special case of a polynomial of degree 2 in the quadratic space $(\mathbb{R}^n,\| \cdot \|_2^2)$ as
%, the Euclidean norm of a vector $x \in \mathbb{R}^n$ can be presented by the polynomial $x^2$. Concretely, for $x \in \Cl^{(1)}(\mathbb{R}^n)$:
\begin{equation}
    \|x\|_2^2 = \mathfrak{q}(x) = (x^2)^{(0)}.
\end{equation}
%Note that, as showed in \cite{satorras2021n}, the geometry of a set of points in $\mathbb{R}^n$ is completely defined by the distances of every pair of points, up to a transformation of $\E(n)$. 
As our experiment results indicated below, CG-EGNNs outperform the existing models since they can flexibly choose to learn their suitable components. %, instead of sticking with a strict one. 

Moreover, CG-EGNNs achieve better performance in most cases when high-order messages are involved. One advantage of our model is, with the Clifford layers, to incorporate high-order messages. We simply aggregate features of neighbors of vertex $i$ and put it through such layers together with features of $i$. Compared to different works from the literature, we have to find a function for more than two position information, that can capture equivariance like the Euclidean norm. This approach is non-intuitive and impractical. More importantly, it is likely to result in the loss of information about the initial position of the system.

\section{Adding $k$-hop Message Passing Mechanism to CG-EGNNs}\label{sec:k-hop}

\subsection{Necessity of Adding the $k$-hop Message Passing to CG-EGNNs}

The Clifford graph convolution layer presented in Section~\ref{sec:methodology} %[\tanr{in Section XXX}] 
has an ability of gaining richer information from neighbors to each node feature update.
However, in some special cases that the graph has only a few edges, such as trees, this ability becomes unclear.
The main reason is that, in this case, each node does not have enough neighbors to compute even one high-order message. Then, the messages $\mathbf{m}^{(d)}$ for some $d > 1$ will disappear.
We can ignore this issue if we only compute high-order messages at nodes that have enough neighbors and leave the high-order messages at the remaining nodes %[\tanr{the other high-order message"}] 
to be none (which is zero, by Eq.~\eqref{computehigherordermessage}). Alternatively, we skip the adjacency matrix and assume the graph is fully connected. Both of these approaches might not scale well to large graphs because there is the risk of losing or overflowing information.
To solve this issue efficiently, for each node $i$, we can consider the exchange of messages at $i$ to a suitably larger set of vertices than the neighborhood of $i$. We will replace the neighborhood $\mathcal{N}(i)$ of $i$ by a larger set, which is the $k-$hop neighborhood for a positive integer $k$, defined as follows:
\begin{equation}
    \mathcal{N}^k(i) = \{ j \in \mathcal{V} ~ | ~ \operatorname{d}_G(i,j) \leq k \text{ and } j \neq i \}.
\end{equation}
In other words, $\mathcal{N}^{k}(i)$ is the set of nodes $j$ that differs from $i$ such that there exists a path of length at most $k$ from $i$ to $j$. Note that $\mathcal{N}^{1}(i) = \mathcal{N}(i)$. Now, for each $i \in \mathcal{V}$ and $d = 1, \ldots, D$, we compute message from each subset $A \subseteq \mathcal{N}^k(i)$ with $d$ elements contributing to the node $i$ as in Eq.~\eqref{partialmessage}:
\begin{equation}
\mathbf{m}_{i,A} = \phi_m^{(d)} \left( \mathbf{h}_i^{l},\sum_{j \in A} \mathbf{h}_j^{l} \right), \forall   A \subseteq \mathcal{N}^k(i), |A| = d, 
\end{equation}
and rewrite Eq.~\eqref{computehigherordermessage} as:
\begin{equation}
    \mathbf{m}_i^{(d)} = \sum_{A \subseteq \mathcal{N}^k(i),\, |A|=d} \mathbf{m}_{i,A},\quad d=1,\ldots,D.
\end{equation}
In our experiments, graphs considered in the n-body system are complete graphs, but the ones in CMU Motion Capture and MD17 are trees or trees with a few additional edges. 
So using the $k-$hop neighborhood in these cases will be reasonable and necessary.

\subsection{Universality of the $k$-hop message passing mechanism in geometric GNNs}

%To avoid not necessarily technical involving when verifying the universality of our model, we will restrict ourself on the zero order components of the Clifford algebras.
%In particular, the input $\mathbf{h}^l=\{\mathbf{h}_1^l,\ldots,\mathbf{h}_M^l\} \subset \text{Cl}(\mathbb{R}^n)^{\text{nf}}$ of the Clifford Graph Convolution Layer is now replaced by a subset $\mathbf{z}=\{(\mathbf{h}^l_1)^{(0)},\ldots,(\mathbf{h}_M)^{(0)}\}$ of $\mathbb{R}^d$ with $d = \text{nf}$.
%In this context, we do not have to care about the $O(n)$-equivariant of the considered graph neural networks anymore as the full Clifford algebras will handle it.
%It is noted that, since the zero components form a subalgebra of the Clifford algebras, the restriction of every polynomial map over the Clifford algebras on their zero components is again a polynomial map.

To verify the universality of $k$-hop message passing mechanism of geometric GNNs, let us forget the Clifford algebraic structure and the high-order component for a while, and consider node features of the geometric graphs as vectors in $\mathbb{R}^d$ as in the original setting.

The universality or the expressive power of the $k$-hop message passing GNNs for generic graphs has been intensively studied in the literature (see for instance in \cite{feng2022powerful}).
However, unlike generic graphs, a geometric graph is a special kind of graph whose nodes are associated with coordinate features in $\mathbb{R}^d$. 
%In addition to the symmetries arising from permuting of vertices, geometric graphs often exhibit other physical symmetries of translations, rotations, and reflections, making them ineffectively processed by current Graph Neural Networks (GNNs). 
Because of the node features, two geometric graphs having the same combinatorial graph structure can be different in the geometric sense.
Therefore, the results of the universality of generic graphs cannot be applied directly to those of geometric graphs.
To tackle this issue, we establish the universality of the $k$-hop message passing mechanism for geometric GNNs.

Let us consider a graph $G$ with $M$ nodes which are points in $\mathbb{R}^d$.
By using normalization if necessary, we can assume that $G$ is a subset of the box $[0,1]^d \subset \mathbb{R}^d$.
According to the limitation of handling small numbers by computers, we can assume that the distance of every two distinct nodes of a considered graph is always greater than a fixed small real number $\alpha>0$.
Let $\mathcal{X}$ be the space of all such graphs defined as
\begin{align}\label{eq:X}
    \mathcal{X} =  \left\{G \subset [0,1]^d \text{ with }|G|=M ~ | ~ \text{ for all } x \neq y \text{ in } G:  \|x-y\|_{\infty} \geq \alpha  \right\}.
\end{align}
%sets containing $M$ distinct points of $[0,1]^d \subset \mathbb{R}^d$.
A graph neural network can now be simply regarded as a parametrized function that maps each graph in $\mathcal{X}$ to a prediction vector in $\mathbb{R}^r$.
In this section, we will prove that our models constructed according to the design in Section~\ref{sec:methodology}, %[\tanr{refer the right section here}], 
but without the context of Clifford algebras, can be used to approximate any continuous function from $\mathcal{X}$ to $\mathbb{R}^r$.

For each pair of graphs $G,G'$ in $\mathcal{X}$, the Hausdorff distance $d_H$ between $G$ and $G'$ is defined as 
\begin{align}
 d_H(G,G') =  \max \Bigl\{ \sup_{z \in G}  \inf_{z' \in G'}\|z-z'\|_{\infty},
	\sup_{z' \in G'} \inf_{z \in G}\|z-z'\|_{\infty} \Bigr\}.
\end{align}
Given this distance function, the space $\mathcal{X}$ becomes a compact metric space \citep{henrikson1999completeness}.
Therefore, we can discuss continuous functions on $\mathcal{X}$ with respect to this metric.
The proof for the following theorem can be found in Appendix~\ref{appendix:universalproof}.

% https://www2.math.upenn.edu/~brweber/Lectures/USTC2011/Lecture%205%20-%20Gromov-Hausdorff%20distance.pdf
%}

\begin{theorem}\label{thm:universal}
Let $f \colon \mathcal{X} \to \mathbb{R}^r$ be a continuous map.
Then for every $\epsilon>0$, there exists a positive integer $N$, continuous functions $\phi_m \colon \mathbb{R}^{d} \to \mathbb{R}^{N}$ and $\phi_h \colon \mathbb{R}^{N} \to \mathbb{R}^r$ such that: 
\begin{align}
\left\Vert f(G)- \phi_h \left( \sum_{z \in G} \phi_m(z) \right) \right\Vert_{\infty} < \epsilon,
\end{align}
for every graph $G$ in $\mathcal{X}$.
\end{theorem}

\begin{remark}[{Universality}]
In Theorem~\ref{thm:universal}, the continuous function $\phi_h(\cdot)$ stands for the aggregation $\phi_h(\mathbf{h}_i^{l},\cdot)$, while the continuous function $\phi_m(\cdot)$ stands for the message $\phi_m(\mathbf{h}_i^{l},\cdot)$ (see Eqs.~\eqref{partialmessage}-\eqref{featurecompute} in the Clifford Graph Convolution Layer).
However, to have the sum $\sum_{z \in G} \phi_m(z)$ in the message passing process, two conditions are required: first, the graph $G$ is connected (which is often the case), and second, the $k$-hop neighborhood $\mathcal{N}^k$ much be large enough to cover all nodes of the graph.
In practice, increasing $k$ will increase the size of the $k$-hop neighborhood $\mathcal{N}^k$ very fast.
However, to achieve satisfactory results in experiments, there is no need to increase $k$ too much.
In addition, the functions $\phi_h$ and $\phi_m$ can be approximated further by MLPs regarding the universal approximation theorem \cite{pinkus1999approximation}.
As a consequence, this theorem asserts the universality of the $k$-hop message passing mechanism.
\end{remark}

\begin{remark}
    In Theorem~\ref{thm:universal}, the composition $\phi_h \circ \sum \circ \phi_m$ is equivariant with respect to the permutation group but not the orthogonal group.
    The Clifford group structures play the role of adding the equivariant property (with respect to the orthogonal group) but still keep the high-order message passing mechanism of the original form to maintain the universality as much as possible. 
\end{remark}

\section{Experimental Results}
\label{sec:experiments}
In the experiment session, we demonstrate that CG-EGNN attains top performance compared to other equivariant models across different benchmarks. For our CG-EGNN models, we denote the variant that incorporates only first-order messages as CG-EGNN-$1$, the variant that incorporates both first-order and second-order messages as CG-EGNN-$1$-$2$, and analogously for other combinations of message orders. The training setups are kept as similar to the other baselines as possible. All the hyperparameter settings and experiment details can be found in Appendix~\ref{appendix:implementation_details}.
% ----
% \begin{table}[!t]
% % \vskip -0.2in
%   \centering
%     \setlength{\tabcolsep}{2pt}
%   \small
%   \caption{Prediction error of N-body experiment. Results averaged across 4 runs. }
%   % \resizebox{0.27\textwidth}{!}{
%     \begin{tabular}{lcc}
%     \toprule
%           Model & MSE ($\downarrow$) \\
%     \midrule
%     GNN   & $0.0107$  \\
%     TFN   &  $0.0244$ \\
%     SE(3)-Tr.   &  $0.0244$ \\
%     Radial Field   &  $0.0104$\\
%     EGNN   &  $0.0070$ \\
%     SEGNN   &  $0.0043$ \\
%     CGENN   &   $\underline{0.0039}$\tiny{$\pm{0.0001}$}  \\
%     \midrule
%     CG-EGNN-$1$ & $\underline{0.0039}$\tiny{$\pm{0.0001}$} \\
%     CG-EGNN-$1$-$2$ & $\mathbf{0.0035}$\tiny{$\pm{0.0001}$} \\
%     CG-EGNN-$1$-$2$-$3$ & $\mathbf{0.0035}$\tiny{$\pm{0.0001}$} \\
%     \bottomrule
%     \end{tabular}%
%     % }
%   \label{tab:nbody}%
%   \vspace{-0.1in}
% \end{table}%
% transpose
\begin{table}[t]
\caption{MSE ($\times 10^{-2}$) of N-body experiment. Results averaged across 4 runs. }
\medskip
\centering

    \begin{adjustbox}{width=1.0\textwidth}
    \begin{tabular}{cccccccccc}
    \toprule
    \multirow{2}{*}{GNN}      & \multirow{2}{*}{TFN}      & \multirow{2}{*}{SE(3)-Tr.} & \multirow{2}{*}{Radial Field} & \multirow{2}{*}{EGNN}     & \multirow{2}{*}{SEGNN}    & \multirow{2}{*}{CGENN}                         & \multicolumn{3}{c}{CG-EGNN}     \\ 
    \cmidrule{8-10}
    &&&&&&& $1$  & $1$-$2$ & $1$-$2$-$3$  \\
    \midrule
$1.07$ & $2.44$ & $2.44$  & $1.04$     & $0.70$ & $0.43$ & $\underline{0.39\pm0.01}$ & $\underline{0.39\pm0.01}$ & $\mathbf{0.35\pm0.01}$ & $\mathbf{0.35\pm0.01}$ \\ 
\bottomrule
    \end{tabular}
    \end{adjustbox}
  \label{tab:nbody}
% \vspace{-0.1}
\vspace{-0.1in}
\end{table}%

\begin{table}[t]
  \caption{MSE ($\times 10^{-2}$) of CMU motion capture with~\citep{huang2022equivariant} settings and~\citep{han2022equivariant} settings. Results averaged across 4 runs.}
    \medskip
    \centering
    \begin{adjustbox}{width=1.0\textwidth}
\begin{tabular}{cccccccccc}
\toprule
Model                                      & GNN              & TFN               & SE(3)-Tr.         & Radial Field       & EGNN              & GMN               & EGHN            & CG-EGNN-$1$               & CG-EGNN-$1$-$2$        \\
\midrule
Settings from~\citep{huang2022equivariant} & $67.3${$\pm1.1$} & $66.9${$\pm 2.7$} & $60.9${$\pm 0.9$} & $197.0${$\pm 1.0$} & $59.1${$\pm 2.1$} & $43.9${$\pm 1.1$} & --              & $\underline{23.2\pm 3.3}$ & $\mathbf{18.0\pm 0.6}$ \\
Settings from~\citep{han2022equivariant}   & $36.1${$\pm1.5$} & $32.0${$\pm1.8$}  & $31.5${$\pm2.1$}  & $188.0${$\pm1.9$}  & $28.7${$\pm1.6$}  & $21.6${$\pm1.5$}  & $8.5${$\pm2.2$} & $\underline{4.9\pm0.6}$   & $\mathbf{4.3\pm0.3}$   \\
\bottomrule
\end{tabular}
\end{adjustbox}
\label{tab:motioncapture_all}
\vspace{-0.2in}
\end{table}
\begin{table}[t]
  \caption{MSE ($\times 10^{-2}$) on MD17 dataset. Results averaged across 4 runs.}  
\medskip\centering
    \begin{adjustbox}{width=1.0\textwidth}
    \begin{tabular}{lcccccccc}
    \toprule
          & Aspirin & Benzene & Ethanol & Malonaldehyde & Naphthalene & Salicylic & Toluene & Uracil \\
    \midrule
    RF    & $10.94${$\pm0.01$}   & $103.72${$\pm1.29$}   & $4.64${$\pm0.01$}   & $13.93${$\pm0.03$}   & $0.50${$\pm0.01$}   & $1.23${$\pm0.01$}   & $10.93${$\pm0.04$}   & $0.64${$\pm0.01$}   \\
    TFN   & $12.37${$\pm0.18$}   & $58.48${$\pm1.98$}   & $4.81${$\pm0.04$}   & $13.62${$\pm0.08$}   & $0.49${$\pm0.01$}   & $1.03${$\pm0.02$}   & $10.89${$\pm0.01$}   & $0.84${$\pm0.02$}   \\
    SE(3)-Tr. & $11.12${$\pm0.06$}   & $68.11${$\pm0.67$}   & $4.74${$\pm0.13$}   & $13.89${$\pm0.02$}   & $0.52${$\pm0.01$}   & $1.13${$\pm0.02$}   & $10.88${$\pm0.06$}   & $0.79${$\pm0.02$}  \\
    EGNN  & $14.41${$\pm0.15$}  & $62.40${$\pm0.53$}  & $4.64${$\pm0.01$}  & $13.64${$\pm0.01$}  & $0.47${$\pm0.02$}  & $1.02${$\pm0.02$}  & $11.78${$\pm0.07$}  & $0.64${$\pm0.01$}  \\
    EGNNReg & $13.82${$\pm0.19$}  & $61.68${$\pm0.37$}  & $6.06${$\pm0.01$}  & $13.49${$\pm0.06$}  & $0.63${$\pm0.01$}  & $1.68${$\pm0.01$}  & $11.05${$\pm0.01$}  & $0.66${$\pm0.01$}  \\

    GMN   & $10.14${$\pm0.03$}  & $48.12${$\pm0.40$}  & $4.83${$\pm0.01$}  & $13.11${$\pm0.03$}  & $0.40${$\pm0.01$}  & $0.91${$\pm0.01$}  & $10.22${$\pm0.08$}  & $0.59${$\pm0.01$}  \\
    
        GMN-L   & $9.76${$\pm0.11$}  & $54.17${$\pm0.69$}  & $\mathbf{4.63\pm0.01}$  & $\mathbf{12.82\pm0.03}$  & $0.41${$\pm0.01$}  & \underline{$0.88\pm0.01$}  & $10.45${$\pm0.04$}  & $0.59${$\pm0.01$}  \\
        \midrule
        CG-EGNN-$1$   & \underline{$9.47\pm0.09$}  & \underline{$38.14\pm0.44$}  & $4.65${$\pm0.01$}  & $\mathbf{12.82\pm0.03}$ & \underline{$0.33\pm0.01$}  & $0.91${$\pm0.04$}  & \underline{$10.13\pm0.04$}  & \underline{$0.55\pm0.01$}  \\
        CG-EGNN-$1$-$2$  & $\mathbf{9.39\pm0.06}$  & $\mathbf{37.45\pm0.30}$  & \underline{$4.64\pm0.01$}  & \underline{$12.84\pm0.05$}  & $\mathbf{0.31\pm0.01}$  & $\mathbf{0.82\pm0.02}$  & $\mathbf{10.11\pm0.04}$  & $\mathbf{0.54\pm0.01}$  \\
        % $\Delta$Bond  & 0.18  & 0.08  & 0.31  & 0.37 & 0.15  & 0.17  & 0.19  & 0.17 \\
    \bottomrule
    \end{tabular}
    \end{adjustbox}
  \label{tab:md17}%
  \vspace{-0.2in}
\end{table}
% ----
\noindent \textbf{N-body System.} We conduct the n-body experiment to measure the performance of our model on the task of simulating physical systems. In this experiment, we simulate the dynamics of $n=5$ charged particles in a 3D space. 
% Each particle has an initial position and velocity parameters and undergoes simulation for $1000$ timesteps. The objective for each model is to predict the displacement of each particle at the final timestep. To ensure translational invariance, we subtract the mean positions from the particles's positions.
%  Additionally, the input to the neural network includes the initial velocity and electrical charge of each particle. The n-body system is treated as a fully connected system, where an edge connects each pair of particles. The edge attributes are the product of charges for connected node pairs.
We train our CG-EGNN networks with 4 different seeds and compare the performance with the following baselines: GNN~\citep{pmlr-v70-gilmer17a}, TFN~\citep{thomas2018tensor}, SE(3)-Transformer~\citep{fuchs2020se}, Radial Field~\citep{kohler2020equivariant}, EGNN~\citep{satorras2021n}, SEGNN~\citep{brandstetter2021geometric}, and CGENN~\citep{ruhe2023clifford}. 
% For this task, CGENN~\citep{ruhe2023clifford} is similar to our $1$-CG-EGNN, so we omit ours and only show $2$-CG-EGNN to demonstrate how adding high-order messages can significantly improve the model's expressivity.

The results presented in Table~\ref{tab:nbody} indicate that CG-EGNN-$1$-$2$ significantly outperforms all other baselines, achieving the lowest Mean Squared Error (MSE) for this task.  
This illustrates that by incorporating high order messages, our CG-EGNN-$1$-$2$ and CG-EGNN-$1$-$2$-$3$ exhibit enhanced capabilities in modeling physical systems compared to the baselines. While incorporating third-order messages does not yield performance gains in this experiment, we argue that higher-order messages become more beneficial as the graph size increases. Since this experiment only involves graphs with 5 nodes, the advantages of higher-order messages may not have been fully realized. We demonstrate the impact of higher-order messages for other tasks with more nodes in the ablation study below.

\noindent \textbf{CMU Motion Capture.} In the first experiment, we keep all settings the same as GMN paper~\citep{huang2022equivariant} and use the sets of sticks and hinges from~\citep{huang2022equivariant} embedded as the edge features. Table~\ref{tab:motioncapture_all} demonstrates that CG-EGNN-$1$ outperforms the baselines EGNN, GMN by a large margin, and adding the second-order messages, CG-EGNN-$1$-$2$ further lowers the MSE. In the second experiment, we adopt the experiment settings detailed in the EGHN paper~\citep{han2022equivariant}: the node feature is augmented by the $z$-axis coordinates, resulting in a model that is height-aware and maintains equivariance in the horizontal directions. Table~\ref{tab:motioncapture_all} also demonstrates that our model persistently outperforms the baseline models under the new distinct settings. Specifically, the MSE of our model is nearly halved compared to EGHN, and only a small fraction of EGNN and GMN.

\noindent \textbf{MD17.} Table~\ref{tab:md17} demonstrated that CG-EGNN-$1$-$2$ attains the lowest MSE for $6$ out of $8$ molecules and our models have the lowest MSE in general. It is noteworthy that CG-EGNN-$1$-$2$ exhibits significantly better performance on more complex, cyclic molecules, namely aspirin, benzene, naphthalene, salicylic, toluene, and uracil while remaining competitive other simpler, open-chain molecules. Figure~\ref{fig:molecules} from Appendix~\ref{appendix:implementation_details} visualizes the clear distinction between the more complex molecules and the simpler molecules in the dataset. This observation serves as a strong indication of our model's ability in capturing high-order information inherent in the molecular graphs.

% ----
% {\color{blue}
% \subsection{5D Convex Hulls}
% \label{subsec:5D_convex_hulls}
% \begin{table}[t]
% % \vskip -0.2in
%   \centering
%     \setlength{\tabcolsep}{2pt}
%   \small
%   \caption{\textcolor{blue}{Prediction error of the 5D Convex Hulls experiment. Results for EMPSN and CSMPN are as reported in~\cite{liu2024clifford}.}}
%   % \resizebox{0.22\textwidth}{!}{
%     \begin{tabular}{lcc}
%     \toprule
%          Model & MSE ($\downarrow$) \\
%     \midrule
%     GNN  & $0.0317$   \\
%     EGNN & $0.0123$ \\
%     CGENN  &  $0.0152$  \\
%     EMPSN  &  $0.0070$ \\
%     CSMPN  &  $\underline{0.0020}$  \\
%     \midrule
%     CG-EGNN-$1$ & $0.0055$ \\
%     CG-EGNN-$1$-$2$ & $\mathbf{0.0009}$ \\
%     \bottomrule
%     \end{tabular}%
%     % }
%   \label{tab:5d_convex_hull}%
%   \vskip -0.2in
% \end{table}
% ----
% transpose
\label{subsec:5D_convex_hulls}

\begin{table}[t]
  \caption{Prediction error of the 5D Convex Hulls experiment. Results for EMPSN and CSMPN are as reported in~\cite{liu2024clifford}.}
    \medskip
    \centering
    \begin{adjustbox}{width=1.0\textwidth}
    \begin{tabular}{cccccccc}
    \toprule
     Model              & GNN      & EGNN     & CGENN    & EMPSN    & CSMPN                & CG-EGNN-$1$ & CG-EGNN-$1$-$2$   \\ \midrule
MSE ($\downarrow$) & $0.0317$ & $0.0123$ & $0.0152$ & $0.0070$ & $\underline{0.0020}$ & $0.0055$    & $\mathbf{0.0009}$\\
    \bottomrule
    \end{tabular}
    \end{adjustbox}
  \label{tab:5d_convex_hull}%
  \vskip -0.2in
\end{table}

% ----
\noindent \textbf{5D Convex Hulls.} To evaluate the performance of the CG-EGNN model on higher-dimensional data, we conduct an experiment estimating the volume of 5D convex hulls, following the methodology outlined in the work of~\cite{liu2024clifford}. Table \ref{tab:5d_convex_hull} demonstrates that our CG-EGNN-$1$-$2$ attains the best performance, with MSE of 0.0009. Notably, the second-best model is CSMPN, which is also a higher-order message passing network. These results provide strong evidence that higher-order message passing architectures can significantly enhance performance on tasks involving higher-dimensional data.

\noindent \textbf{Ablation study on the effect of including high order messages on 3D convex hull dataset.}
\label{subsec:ablation_high_order}
We perform an ablation study on the effect of each combination of high-order messages up to order 3 on the performance of our CG-EGNN models. For this task, we run 3 sub-experiments estimating the volume of a 3D convex hull with the number of nodes per graph $\in \{6, 7, 8\}$.

The results reported in Table~\ref{tab:ablation_study} empirically validate the significant performance gains achieved by the CG-EGNN model when incorporating higher-order message passing. The CG-EGNN-$2$ variant has the best performance among the models considering one message order. Furthermore, the CG-EGNN-$1$-$2$ model is the top-performing architecture when two message orders are included. Additionally, the CG-EGNN-$1$-$2$-$3$ variant, encompassing three message orders, outperforms all other models. These findings provide strong empirical evidence that including more higher order messages enables more effective learning of intricate geometric representations, thereby enhancing the model's ability to capture complex structural patterns.
Additionally, all model runtime are reported in Table~\ref{tab:ablation_study_runtime} from Appendix~\ref{appendix:ablation_study}. Consistent with previous research on using Clifford algebra to construct neural networks, our models exhibit higher time complexity compared to other models. Enhancing the time complexity of Clifford GNNs by optimizing the implementation of Clifford algebra operators remains an open challenge for future work.

\section{Concluding Remarks}

We introduced a novel $\E(n)$-equivariant graph neural network that incorporates a high-order message passing mechanism within the framework of Clifford algebras. 
Unlike previous equivariant graph neural networks, our model has the ability to learn its favorite equivariant functions from the positional features of data points, thereby extracting more comprehensive information from neighboring nodes during the message passing process. 
This enhanced capability enables our model to capture equivariance effectively, while harnessing the expressive power inherent in high-order message passing mechanisms. 
We believe that these properties make our approach highly effective in geometric deep learning promising in various applications in the fields of chemistry and physical sciences.
A limitation of our method, as well as other existing models built upon Clifford algebras, is the increased computational cost of the high-order message passing mechanism. However, we have already made significant improvements in experiments by adjusting key components of GNNs while maintaining number of parameters, so we are optimistic about future advancements in this direction.

\newpage

\bibliography{example_paper}

\begin{thebibliography}{10}

\bibitem{ablamowicz2004lectures}
Rafal Ablamowicz, Garret Sobczyk, et~al.
\newblock {\em Lectures on Clifford (geometric) algebras and applications}.
\newblock Springer, 2004.

\bibitem{alex2011numerical}
Arne Alex, Matthias Kalus, Alan Huckleberry, and Jan von Delft.
\newblock A numerical algorithm for the explicit calculation of {SU(N)} and
  {SL(N,C)} {Clebsch-Gordan} coefficients.
\newblock {\em Journal of Mathematical Physics}, 52(2), 2011.

\bibitem{anderson2019cormorant}
Brandon Anderson, Truong~Son Hy, and Risi Kondor.
\newblock Cormorant: Covariant molecular neural networks.
\newblock {\em Advances in neural information processing systems}, 32, 2019.

\bibitem{batatia2022design}
Ilyes Batatia, Simon Batzner, D{\'a}vid~P{\'e}ter Kov{\'a}cs, Albert Musaelian,
  Gregor~NC Simm, Ralf Drautz, Christoph Ortner, Boris Kozinsky, and G{\'a}bor
  Cs{\'a}nyi.
\newblock The design space of e (3)-equivariant atom-centered interatomic
  potentials.
\newblock {\em arXiv preprint arXiv:2205.06643}, 2022.

\bibitem{batatia2024general}
Ilyes Batatia, Mario Geiger, Jose Munoz, Tess Smidt, Lior Silberman, and
  Christoph Ortner.
\newblock A general framework for equivariant neural networks on reductive lie
  groups.
\newblock {\em Advances in Neural Information Processing Systems}, 36, 2024.

\bibitem{batatia2022mace}
Ilyes Batatia, David~P Kovacs, Gregor Simm, Christoph Ortner, and G{\'a}bor
  Cs{\'a}nyi.
\newblock Mace: Higher order equivariant message passing neural networks for
  fast and accurate force fields.
\newblock {\em Advances in Neural Information Processing Systems},
  35:11423--11436, 2022.

\bibitem{batzner20223}
Simon Batzner, Albert Musaelian, Lixin Sun, Mario Geiger, Jonathan~P Mailoa,
  Mordechai Kornbluth, Nicola Molinari, Tess~E Smidt, and Boris Kozinsky.
\newblock E (3)-equivariant graph neural networks for data-efficient and
  accurate interatomic potentials.
\newblock {\em Nature communications}, 13(1):2453, 2022.

\bibitem{bayro2006conformal}
Eduardo Bayro-Corrochano, Leo Reyes-Lozano, and Julio Zamora-Esquivel.
\newblock Conformal geometric algebra for robotic vision.
\newblock {\em Journal of Mathematical Imaging and Vision}, 24:55--81, 2006.

\bibitem{bekkers2019b}
Erik~J Bekkers.
\newblock B-spline cnns on lie groups.
\newblock {\em arXiv preprint arXiv:1909.12057}, 2019.

\bibitem{bhatti2021advanced}
Uzair~Aslam Bhatti, Zhou Ming-Quan, Huo Qing-Song, Sajid Ali, Aamir Hussain,
  Yan Yuhuan, Zhaoyuan Yu, Linwang Yuan, and Saqib~Ali Nawaz.
\newblock Advanced color edge detection using clifford algebra in satellite
  images.
\newblock {\em IEEE Photonics Journal}, 13(2):1--20, 2021.

\bibitem{bochkarev2022multilayer}
Anton Bochkarev, Yury Lysogorskiy, Christoph Ortner, G{\'a}bor Cs{\'a}nyi, and
  Ralf Drautz.
\newblock Multilayer atomic cluster expansion for semilocal interactions.
\newblock {\em Physical Review Research}, 4(4):L042019, 2022.

\bibitem{brandstetter2021geometric}
Johannes Brandstetter, Rob Hesselink, Elise van~der Pol, Erik~J Bekkers, and
  Max Welling.
\newblock Geometric and physical quantities improve e (3) equivariant message
  passing.
\newblock {\em arXiv preprint arXiv:2110.02905}, 2021.

\bibitem{brehmer2023geometric}
Johann Brehmer, Pim De~Haan, S{\"o}nke Behrends, and Taco Cohen.
\newblock Geometric algebra transformers.
\newblock {\em arXiv preprint arXiv:2305.18415}, 2023.

\bibitem{breuils2022new}
Stephane Breuils, Kanta Tachibana, and Eckhard Hitzer.
\newblock New applications of clifford’s geometric algebra.
\newblock {\em Advances in Applied Clifford Algebras}, 32(2):17, 2022.

\bibitem{bruna2013spectral}
Joan Bruna, Wojciech Zaremba, Arthur Szlam, and Yann LeCun.
\newblock Spectral networks and locally connected networks on graphs.
\newblock {\em arXiv preprint arXiv:1312.6203}, 2013.

\bibitem{chmiela2017machine}
Stefan Chmiela, Alexandre Tkatchenko, Huziel~E Sauceda, Igor Poltavsky,
  Kristof~T Sch{\"u}tt, and Klaus-Robert M{\"u}ller.
\newblock Machine learning of accurate energy-conserving molecular force
  fields.
\newblock {\em Science advances}, 3(5):e1603015, 2017.

\bibitem{cmu2003motion}
CMU.
\newblock Carnegie-mellon motion capture database, 2003.

\bibitem{cohen2016group}
Taco Cohen and Max Welling.
\newblock Group equivariant convolutional networks.
\newblock In {\em International conference on machine learning}, pages
  2990--2999. PMLR, 2016.

\bibitem{coors2018spherenet}
Benjamin Coors, Alexandru~Paul Condurache, and Andreas Geiger.
\newblock Spherenet: Learning spherical representations for detection and
  classification in omnidirectional images.
\newblock In {\em Proceedings of the European conference on computer vision
  (ECCV)}, pages 518--533, 2018.

\bibitem{defferrard2016convolutional}
Micha{\"e}l Defferrard, Xavier Bresson, and Pierre Vandergheynst.
\newblock Convolutional neural networks on graphs with fast localized spectral
  filtering.
\newblock {\em Advances in neural information processing systems}, 29, 2016.

\bibitem{deng2021vector}
Congyue Deng, Or~Litany, Yueqi Duan, Adrien Poulenard, Andrea Tagliasacchi, and
  Leonidas~J Guibas.
\newblock Vector neurons: A general framework for {SO(3)}-equivariant networks.
\newblock In {\em Proceedings of the IEEE/CVF International Conference on
  Computer Vision}, pages 12200--12209, 2021.

\bibitem{eijkelboom2023n}
Floor Eijkelboom, Rob Hesselink, and Erik~J Bekkers.
\newblock E $(n) $ equivariant message passing simplicial networks.
\newblock In {\em International Conference on Machine Learning}, pages
  9071--9081. PMLR, 2023.

\bibitem{feng2022powerful}
Jiarui Feng, Yixin Chen, Fuhai Li, Anindya Sarkar, and Muhan Zhang.
\newblock How powerful are k-hop message passing graph neural networks.
\newblock {\em Advances in Neural Information Processing Systems},
  35:4776--4790, 2022.

\bibitem{finzi2020generalizing}
Marc Finzi, Samuel Stanton, Pavel Izmailov, and Andrew~Gordon Wilson.
\newblock Generalizing convolutional neural networks for equivariance to lie
  groups on arbitrary continuous data.
\newblock In {\em International Conference on Machine Learning}, pages
  3165--3176. PMLR, 2020.

\bibitem{fuchs2020se}
Fabian Fuchs, Daniel Worrall, Volker Fischer, and Max Welling.
\newblock Se (3)-transformers: 3d roto-translation equivariant attention
  networks.
\newblock {\em Advances in neural information processing systems},
  33:1970--1981, 2020.

\bibitem{Garling_2011}
D.~J.~H. Garling.
\newblock {\em Clifford Algebras: An Introduction}.
\newblock London Mathematical Society Student Texts. Cambridge University
  Press, 2011.

\bibitem{gasteiger2021gemnet}
Johannes Gasteiger, Florian Becker, and Stephan G{\"u}nnemann.
\newblock Gemnet: Universal directional graph neural networks for molecules.
\newblock {\em Advances in Neural Information Processing Systems},
  34:6790--6802, 2021.

\bibitem{gasteiger2020directional}
Johannes Gasteiger, Janek Gro{\ss}, and Stephan G{\"u}nnemann.
\newblock Directional message passing for molecular graphs.
\newblock {\em arXiv preprint arXiv:2003.03123}, 2020.

\bibitem{gilmer2017neural}
Justin Gilmer, Samuel~S Schoenholz, Patrick~F Riley, Oriol Vinyals, and
  George~E Dahl.
\newblock Neural message passing for quantum chemistry.
\newblock In {\em International conference on machine learning}, pages
  1263--1272. PMLR, 2017.

\bibitem{pmlr-v70-gilmer17a}
Justin Gilmer, Samuel~S. Schoenholz, Patrick~F. Riley, Oriol Vinyals, and
  George~E. Dahl.
\newblock Neural message passing for quantum chemistry.
\newblock In Doina Precup and Yee~Whye Teh, editors, {\em Proceedings of the
  34th International Conference on Machine Learning}, volume~70 of {\em
  Proceedings of Machine Learning Research}, pages 1263--1272. PMLR, 06--11 Aug
  2017.

\bibitem{grassmann1862ausdehnungslehre}
Hermann Grassmann.
\newblock {\em Die Ausdehnungslehre}, volume~1.
\newblock Enslin, 1862.

\bibitem{hamilton1866elements}
William~Rowan Hamilton.
\newblock {\em Elements of quaternions}.
\newblock Longmans, Green, \& Company, 1866.

\bibitem{han2022equivariant}
Jiaqi Han, Wenbing Huang, Tingyang Xu, and Yu~Rong.
\newblock Equivariant graph hierarchy-based neural networks.
\newblock {\em Advances in Neural Information Processing Systems},
  35:9176--9187, 2022.

\bibitem{han2022geometrically}
Jiaqi Han, Yu~Rong, Tingyang Xu, and Wenbing Huang.
\newblock Geometrically equivariant graph neural networks: A survey.
\newblock {\em arXiv preprint arXiv:2202.07230}, 2022.

\bibitem{henrikson1999completeness}
Jeff Henrikson.
\newblock Completeness and total boundedness of the hausdorff metric.
\newblock {\em MIT Undergraduate Journal of Mathematics}, 1(69-80):10, 1999.

\bibitem{hildenbrand2008inverse}
Dietmar Hildenbrand, Julio Zamora, and Eduardo Bayro-Corrochano.
\newblock Inverse kinematics computation in computer graphics and robotics
  using conformal geometric algebra.
\newblock {\em Advances in applied Clifford algebras}, 18:699--713, 2008.

\bibitem{hitzer2010interactive}
Eckhard Hitzer and Christian Perwass.
\newblock Interactive 3d space group visualization with clucalc and the
  clifford geometric algebra description of space groups.
\newblock {\em Advances in applied Clifford algebras}, 20:631--658, 2010.

\bibitem{hodapp2023equivariant}
Max Hodapp and Alexander Shapeev.
\newblock Equivariant tensor networks.
\newblock {\em arXiv preprint arXiv:2304.08226}, 2023.

\bibitem{huang2022equivariant}
Wenbing Huang, Jiaqi Han, Yu~Rong, Tingyang Xu, Fuchun Sun, and Junzhou Huang.
\newblock Equivariant graph mechanics networks with constraints.
\newblock In {\em International Conference on Learning Representations}, 2022.

\bibitem{jing2020learning}
Bowen Jing, Stephan Eismann, Patricia Suriana, Raphael~JL Townshend, and Ron
  Dror.
\newblock Learning from protein structure with geometric vector perceptrons.
\newblock {\em arXiv preprint arXiv:2009.01411}, 2020.

\bibitem{kipf2016semi}
Thomas~N Kipf and Max Welling.
\newblock Semi-supervised classification with graph convolutional networks.
\newblock {\em arXiv preprint arXiv:1609.02907}, 2016.

\bibitem{klicpera2021gemnet}
Johannes Klicpera, Florian Becker, and Stephan G{\"u}nnemann.
\newblock Gemnet: Universal directional graph neural networks for molecules.
\newblock {\em arXiv e-prints}, pages arXiv--2106, 2021.

\bibitem{kohler2019equivariant}
Jonas K{\"o}hler, Leon Klein, and Frank No{\'e}.
\newblock Equivariant flows: sampling configurations for multi-body systems
  with symmetric energies.
\newblock {\em arXiv preprint arXiv:1910.00753}, 2019.

\bibitem{kohler2020equivariant}
Jonas K{\"o}hler, Leon Klein, and Frank No{\'e}.
\newblock Equivariant flows: exact likelihood generative learning for symmetric
  densities.
\newblock In {\em International conference on machine learning}, pages
  5361--5370. PMLR, 2020.

\bibitem{kondor2018generalization}
Risi Kondor and Shubhendu Trivedi.
\newblock On the generalization of equivariance and convolution in neural
  networks to the action of compact groups.
\newblock In {\em International Conference on Machine Learning}, pages
  2747--2755. PMLR, 2018.

\bibitem{liu2024clifford}
Cong Liu, David Ruhe, Floor Eijkelboom, and Patrick Forré.
\newblock Clifford group equivariant simplicial message passing networks.
\newblock {\em International Conference on Learning Representations}, 2024.

\bibitem{loshchilov2016sgdr}
Ilya Loshchilov and Frank Hutter.
\newblock Sgdr: Stochastic gradient descent with warm restarts.
\newblock {\em arXiv preprint arXiv:1608.03983}, 2016.

\bibitem{maron2019provably}
Haggai Maron, Heli Ben-Hamu, Hadar Serviansky, and Yaron Lipman.
\newblock Provably powerful graph networks.
\newblock {\em Advances in neural information processing systems}, 32, 2019.

\bibitem{maron2018invariant}
Haggai Maron, Heli Ben-Hamu, Nadav Shamir, and Yaron Lipman.
\newblock Invariant and equivariant graph networks.
\newblock {\em Proceedings of the AAAI conference on artificial intelligence},
  33(01):4602--4609, 2018.

\bibitem{maron2019universality}
Haggai Maron, Ethan Fetaya, Nimrod Segol, and Yaron Lipman.
\newblock On the universality of invariant networks.
\newblock In {\em International conference on machine learning}, pages
  4363--4371. PMLR, 2019.

\bibitem{melnyk2021embed}
Pavlo Melnyk, Michael Felsberg, and M{\aa}rten Wadenb{\"a}ck.
\newblock Embed me if you can: A geometric perceptron.
\newblock In {\em Proceedings of the IEEE/CVF International Conference on
  Computer Vision}, pages 1276--1284, 2021.

\bibitem{morris2019weisfeiler}
Christopher Morris, Martin Ritzert, Matthias Fey, William~L Hamilton, Jan~Eric
  Lenssen, Gaurav Rattan, and Martin Grohe.
\newblock Weisfeiler and leman go neural: Higher-order graph neural networks.
\newblock In {\em Proceedings of the AAAI conference on artificial
  intelligence}, volume~33, pages 4602--4609, 2019.

\bibitem{musaelian2023learning}
Albert Musaelian, Simon Batzner, Anders Johansson, Lixin Sun, Cameron~J Owen,
  Mordechai Kornbluth, and Boris Kozinsky.
\newblock Learning local equivariant representations for large-scale atomistic
  dynamics.
\newblock {\em Nature Communications}, 14(1):579, 2023.

\bibitem{nigam2022unified}
Jigyasa Nigam, Sergey Pozdnyakov, Guillaume Fraux, and Michele Ceriotti.
\newblock Unified theory of atom-centered representations and message-passing
  machine-learning schemes.
\newblock {\em The Journal of Chemical Physics}, 156(20), 2022.

\bibitem{pinkus1999approximation}
Allan Pinkus.
\newblock Approximation theory of the mlp model in neural networks.
\newblock {\em Acta numerica}, 8:143--195, 1999.

\bibitem{pozdnyakov2022incompleteness}
Sergey~N Pozdnyakov and Michele Ceriotti.
\newblock Incompleteness of graph neural networks for points clouds in three
  dimensions.
\newblock {\em Machine Learning: Science and Technology}, 3(4):045020, 2022.

\bibitem{ruhe2023clifford}
David Ruhe, Johannes Brandstetter, and Patrick Forr{\'e}.
\newblock Clifford group equivariant neural networks.
\newblock In {\em Thirty-seventh Conference on Neural Information Processing
  Systems}, 2023.

\bibitem{satorras2021n}
V{\i}ctor~Garcia Satorras, Emiel Hoogeboom, and Max Welling.
\newblock E (n) equivariant graph neural networks.
\newblock In {\em International conference on machine learning}, pages
  9323--9332. PMLR, 2021.

\bibitem{schutt2017schnet}
Kristof Sch{\"u}tt, Pieter-Jan Kindermans, Huziel~Enoc Sauceda~Felix, Stefan
  Chmiela, Alexandre Tkatchenko, and Klaus-Robert M{\"u}ller.
\newblock Schnet: A continuous-filter convolutional neural network for modeling
  quantum interactions.
\newblock {\em Advances in neural information processing systems}, 30, 2017.

\bibitem{schutt2021equivariant}
Kristof Sch{\"u}tt, Oliver Unke, and Michael Gastegger.
\newblock Equivariant message passing for the prediction of tensorial
  properties and molecular spectra.
\newblock In {\em International Conference on Machine Learning}, pages
  9377--9388. PMLR, 2021.

\bibitem{schutt2018schnet}
Kristof~T Sch{\"u}tt, Huziel~E Sauceda, P-J Kindermans, Alexandre Tkatchenko,
  and K-R M{\"u}ller.
\newblock Schnet--a deep learning architecture for molecules and materials.
\newblock {\em The Journal of Chemical Physics}, 148(24), 2018.

\bibitem{shen2024rotation}
Wen Shen, Zhihua Wei, Qihan Ren, Binbin Zhang, Shikun Huang, Jiaqi Fan, and
  Quanshi Zhang.
\newblock Rotation-equivariant quaternion neural networks for 3d point cloud
  processing.
\newblock {\em IEEE Transactions on Pattern Analysis and Machine Intelligence},
  2024.

\bibitem{shen20203d}
Wen Shen, Binbin Zhang, Shikun Huang, Zhihua Wei, and Quanshi Zhang.
\newblock 3d-rotation-equivariant quaternion neural networks.
\newblock In {\em Computer Vision--ECCV 2020: 16th European Conference,
  Glasgow, UK, August 23--28, 2020, Proceedings, Part XX 16}, pages 531--547.
  Springer, 2020.

\bibitem{spellings2021geometric}
Matthew Spellings.
\newblock Geometric algebra attention networks for small point clouds.
\newblock {\em arXiv preprint arXiv:2110.02393}, 2021.

\bibitem{tholke2022torchmd}
Philipp Th{\"o}lke and Gianni De~Fabritiis.
\newblock Torchmd-net: equivariant transformers for neural network based
  molecular potentials.
\newblock {\em arXiv preprint arXiv:2202.02541}, 2022.

\bibitem{thomas2018tensor}
Nathaniel Thomas, Tess Smidt, Steven Kearnes, Lusann Yang, Li~Li, Kai Kohlhoff,
  and Patrick Riley.
\newblock Tensor field networks: Rotation-and translation-equivariant neural
  networks for 3d point clouds.
\newblock {\em arXiv preprint arXiv:1802.08219}, 2018.

\bibitem{trang20243}
Thuan~Anh Trang, Nhat~Khang Ngo, Daniel~T Levy, Thieu~Ngoc Vo, Siamak
  Ravanbakhsh, and Truong~Son Hy.
\newblock E (3)-equivariant mesh neural networks.
\newblock In {\em International Conference on Artificial Intelligence and
  Statistics}, pages 748--756. PMLR, 2024.

\bibitem{wareham2004applications}
Rich Wareham, Jonathan Cameron, and Joan Lasenby.
\newblock Applications of conformal geometric algebra in computer vision and
  graphics.
\newblock In {\em International Workshop on Mathematics Mechanization}, pages
  329--349. Springer, 2004.

\bibitem{xu2018powerful}
Keyulu Xu, Weihua Hu, Jure Leskovec, and Stefanie Jegelka.
\newblock How powerful are graph neural networks?
\newblock {\em arXiv preprint arXiv:1810.00826}, 2018.

\bibitem{zhao2020quaternion}
Yongheng Zhao, Tolga Birdal, Jan~Eric Lenssen, Emanuele Menegatti, Leonidas
  Guibas, and Federico Tombari.
\newblock Quaternion equivariant capsule networks for 3d point clouds.
\newblock In {\em European conference on computer vision}, pages 1--19.
  Springer, 2020.

\end{thebibliography}
\bibliographystyle{plain}

% \begin{ack}
% Use unnumbered first level headings for the acknowledgments. All acknowledgments
% More information about this disclosure can be found at: \url{https://neurips.cc/Conferences/2024/PaperInformation/FundingDisclosure}.

% Do {\bf not} include this section in the anonymized submission, only in the final paper. You can use the \texttt{ack} environment provided in the style file to automatically hide this section in the anonymized submission. 
% \end{ack}

%%%%%%%%%%%%%%%%%%%%%%%%%%%%%%%%%%%%%%%%%%%%%%%%%%%%%%%%%%%%
\newpage
\appendix

\begin{center}
{\bf \Large{Supplement to ``Monomial Matrix Group Equivariant \\ Neural Functional Networks''}}
\end{center}

%\DoToC
\section{Introduction of Clifford Group and Clifford Group Equivariant Neural Networks}

For formal construction and details, see Appendix.~\ref{appendix:cliffordalgebra}, \cite{Garling_2011} and \cite{ruhe2023clifford}.

\subsection{Clifford Algebra}
Let $(V,\mathfrak{q})$ be an $n$-dimensional quadratic space over a field $\mathbb{F}$ with $\operatorname{char} \mathbb{F} \neq 2$. 
The Clifford Algebra, denoted by $\Cl(V,\mathfrak{q})$, is the $\mathbb{F}-$algebra generated by $V$ with relations $v^2 = \mathfrak{q}(v)$ for all $v \in V$, i.e. every element of $\Cl(V,\mathfrak{q})$ is a linear combination of formal products of vectors in $V$ modulo that relation: For all $x \in \Cl(V,\mathfrak{q})$:
\begin{equation}
\label{representation:elementsofClifford}
    x = \sum_{i \in I} c_i\cdot v_{i,1} \cdots v_{i,k_i},
\end{equation}
where $I$ is finite, $c \in \mathbb{F}, v_{i,j} \in V$.
$\Cl(V,\mathfrak{q})$ is an $2^n$-dimensional vector space and it has a decomposition into $n+1$ subspace $\Cl^{(m)}(V,\mathfrak{q})$, $m = 0, \ldots, n$, called grades:
\begin{equation}
    \Cl(V,\mathfrak{q}) = \bigoplus^n_{m=0} \Cl^{(m)}(V,\mathfrak{q})
\end{equation}
We have $\dim_{\mathbb{F}}(\Cl^{(m)}(V,\mathfrak{q})) = \binom{n}{m}$. The field $\mathbb{F}$ and the space $V$ can be identified as $\Cl^{(0)}(V,\mathfrak{q})$ and $\Cl^{(1)}(V,\mathfrak{q})$, respectively. Denote the parity decomposition of $x \in  \Cl(V,\mathfrak{q})$ as $x = x^{[0]} + x^{[1]}$, where $x^{[i]} \in \bigoplus^{m \equiv i \pmod{2}}_{0 \le m \le n} \Cl^{(m)}(V,\mathfrak{q})$ $= \Cl^{[i]}(V,\mathfrak{q})$ for $i =0, 1$.

Let $\Cl^{\times}(V,\mathfrak{q})$ denote the group of invertible elements of $\Cl(V,\mathfrak{q})$. Each $w \in \Cl^{\times}(V,\mathfrak{q})$ defines an endomorphism of $\Cl(V,\mathfrak{q})$ via the (adjusted) twisted conjugation:
\begin{align}
\label{define:rho}
    \rho(w) \colon \Cl(V,\mathfrak{q}) & \longrightarrow \Cl(V,\mathfrak{q}) \notag \\
    x \quad & \longmapsto wx^{[0]}w^{-1} + \alpha(w)x^{[1]}w^{-1},
\end{align}
where $\alpha$ is the main involution of $\Cl(V,\mathfrak{q})$, which is given by $\alpha(w) \coloneqq x^{[0]} - x^{[1]}$.
The Clifford group, denoted by $\Gamma(V,\mathfrak{q})$, is a subgroup of $\Cl^{\times}(V,\mathfrak{q})$ consists of elements that is parity homogeneous and preserves $V$ via $\rho$:
\begin{align}
    \Gamma(V,\mathfrak{q}) = \Bigl\{ w \in  \Cl^{\times}(V,\mathfrak{q}) \cap \left ( \Cl^{[0]}(V,\mathfrak{q}) \cup  \Cl^{[1]}(V,\mathfrak{q}) \right)  | ~   \rho(V) \subset V \Bigr\}.
\end{align}
We can show that for $w \in \Gamma(V,\mathfrak{q})$, $\rho(w)$ is an automorphism of $\Cl(V,\mathfrak{q})$ and preserves each subspace $\Cl^{(m)}(V,\mathfrak{q})$.
This means $\Cl(V,\mathfrak{q})$ and $\Cl^{(m)}(V,\mathfrak{q})$ are group representations of $\Gamma(V,\mathfrak{q})$ (via $\rho$). 
On other hand, the orthogonal group of $(V,\mathfrak{q})$ consists of linear automorphisms of $V$ that preserve quadratic form $\mathfrak{q}$, which is:
\begin{align}
    \Ortho(V,\mathfrak{q}) = \Bigl \{  \text{linear}  \text{ automorphism } f \text{ of } V ~ | ~ \forall v \in V, \mathfrak{q}(f(v)) = \mathfrak{q}(v) \Bigr \}.
\end{align}

Assume that $(V,\mathfrak{q})$ is non-degenerate. By a theorem of Cartan-Dieudonné about every orthogonal transformation in an $n$-dimensional symmetric bilinear space is composition of at most $n$ reflections, we can show that $\Ortho(V, \mathfrak{q})$ can be identified as the Clifford group $\Gamma(V,\mathfrak{q})$ up to multiplication by a invertible scalar. Formally, we have an exact sequence (note that, by \ref{define:rho}, $\rho(w) = \operatorname{id}_{\Cl(V,\mathfrak{q})}$ for all $w \in \mathbb{F}^{\times}$):
\begin{equation}
\label{eq:shortexactsequence}
    \begin{tikzcd}
            0 \arrow{r} & \mathbb{F}^{\times} \arrow[hookrightarrow]{r}{i} & \Gamma(V,\mathfrak{q}) \arrow{r}{\rho|_{V}}  & \OO(V,\mathfrak{q}) \arrow{r}   & 0.
    \end{tikzcd}
\end{equation}

For $w \in \Gamma(V,\mathfrak{q})$, we have $f = \rho|_{V} \in \Ortho(V,\mathfrak{q})$ defines an automorphism of $\Cl(V,\mathfrak{q})$ by $\rho(w)$. Consider $x$ in Eq.~\eqref{representation:elementsofClifford}:
\begin{align}
\label{eq:thesameaction}
    \rho(w)(x) & = \sum_{i \in I} c_i \cdot \rho(w)(v_{i,1}) \cdots \rho(w)(v_{i,k_i}) \notag \\ & = \sum_{i \in I} c_i \cdot f(v_{i,1}) \cdots f(v_{i,k_i}).
\end{align}
So $\rho$ also defines a representation of $\Ortho(V,\mathfrak{q})$ on $\Cl(V,\mathfrak{q})$ which is identical with $\Gamma(V,\mathfrak{q})$ up to multiplication of invertible scalar.
Its subrepresentation on $\Cl^{(0)}(V,\mathfrak{q})$ and $\Cl^{(1)}(V,\mathfrak{q})$ is identity and the canonical representation of $\Ortho(V,\mathfrak{q})$ on $V$.

\subsection{Clifford Group Equivariant Neural Networks (CGENNs)}

In context of equivariance, one important result in \cite{ruhe2023clifford} is grade projections and polynomials with coefficients in $\mathbb{F}$ are $\Gamma(V,\mathfrak{q})$-equivariant. 
\cite{ruhe2023clifford} also provides some $\Gamma(V,\mathfrak{q})$-equivariant layers constructed by these maps, which are linear layer, (fully-connected) geometric product layer, normalization layer and nonlinear activation. 
Details of these layers can be found in Appendix~\ref{appendix:cliffordlayer}. 
Using those layers, we can design $\Gamma(V,\mathfrak{q})$-equivariant neural networks, which we will call Clifford networks. 
Finally, we can optionally embed the input or take grade projection of output to induce neural networks that are $\Ortho(n)$-equivariant.

For the rest of this paper, denote $\Cl(\mathbb{R}^n)$ as the Clifford algebra of the $n$-dimensional real vector space $\mathbb{R}^n$ with quadratic form $\mathfrak{q}$ is the square of the Euclidean norm, i.e.  $\mathfrak{q}(\cdot) = \| \cdot \|_2^2$. In this case, the orthogonal group of $\Ortho(\mathbb{R}^n,\| \cdot \|_2^2)$ is the orthogonal group $\Ortho(n)$:
\begin{equation}
    \Ortho(n) = \{ Q \in \GL(n) ~ | ~ Q^{\top}Q = QQ^{\top} = I_n \}.
\end{equation}
By identifying $\Ortho(n) \simeq \Gamma(\mathbb{R}^n,\| \cdot \|_2^2)  / \mathbb{R}^{\times}$ from Eq.~\eqref{eq:shortexactsequence}, we say that $\Ortho(n)$ acts on $\Cl(\mathbb{R}^n)$ by Eq.~\eqref{eq:thesameaction}.

\section{Clifford Algebra}
\label{appendix:cliffordalgebra}
We follow \cite{ruhe2023clifford} and \cite{Garling_2011}.
Let $\mathbb{F}$ denote a field with $\operatorname{char} \mathbb{F} \neq 2$. Let $V$ be a vector space over $\mathbb{F}$ of finite dimension
$\dim_{\mathbb{F}} V = n$.

\subsection{Clifford Algebra}
\begin{definition}[Quadratic forms and quadratic vector spaces]
A map $\mathfrak{q} \colon V \rightarrow \mathbb{F}$ is called a \textit{quadratic form} of $V$ if for all $c \in \mathbb{F}$ and $v, v_1, v_2 \in V$, we have:
\begin{equation}
    \mathfrak{q}(c \cdot v) = c^2 \cdot \mathfrak{q}(v).
\end{equation}
In this case, the map $\mathfrak{b} \colon V \times V \rightarrow \mathbb{F}$, defined by:
\begin{equation}
    \mathfrak{b}(v_1,v_2) \coloneqq \frac{1}{2} \left ( \mathfrak{q}(v_1+v_2) - \mathfrak{q}(v_1) - \mathfrak{q}(v_2) \right),
\end{equation}
is a bilinear form over $\mathbb{F}$.
The tuple $(V, \mathfrak{q})$ will be called a \textit{quadratic space}. We have a bijective correspondence between \textit{quadratic forms} and \textit{symmetric bilinear forms} on $V$.
\end{definition}

\begin{definition}[Orthogonal basis]
    A basis $e_1, \ldots ,e_n$ of $V$ is called an \textit{orthogonal basis} of $V$ if for all $i \neq j$ we have:
    \begin{equation}
        \mathfrak{b}(e_i,e_j) = 0.
    \end{equation}
    It is called an \textit{orthonormal basis} if, in addition, $\mathfrak{q}(e_i) \in \{-1,0,1\}$ for all $i = 1, \ldots, n.$
\end{definition}

\begin{definition}[Clifford algebra]
    Define the \textit{Clifford algebra} $\Cl(V,\mathfrak{q})$ as the quotient of the tensor algebra of $V$:
    \begin{equation}
        \T(V) \coloneqq \bigoplus^{\infty}_{m=0} V^{\otimes m} = \bigoplus^{\infty}_{m=0} \operatorname{span} \{v_1 \otimes \cdots \otimes v_m ~ | ~ v_1, \ldots, v_m \in V \},
    \end{equation}
    by the ideal
    \begin{equation}
        I(V,\mathfrak{q}) \coloneqq \left < v \otimes v - \mathfrak{q} \cdot 1_{\T(V)} ~ | ~ v \in V \right >,
    \end{equation}
    which is $\Cl(V,\mathfrak{q}) \coloneqq \T(V) / I(V,\mathfrak{q})$. Denote the canonical quotient map as $\pi \colon  \T(V) \rightarrow \Cl(V,\mathfrak{q})$.
\end{definition}

   The quadratic form $\mathfrak{q}$ and and the bilinear form $\mathfrak{b}$ can be canonically extended to $\Cl(V,\mathfrak{q})$, so we can define the orthogonality on $\Cl(V,\mathfrak{q})$.

\begin{theorem}[Basis of Clifford algebra]
    If $e_1,\ldots, e_n$ is any basis of $(V,\mathfrak{q})$ then $(e_A)_{A \subseteq [n]}$ is a basis for $\Cl(V,\mathfrak{q})$, where for $A \subseteq [n]$:
    \begin{equation}
        e_A \coloneqq \prod^{<}_{i \in A} e_i,
    \end{equation}
    where the product is taken in increasing order of the indices $i \in A$. 

    In other words, $\dim_{\Rbb} \Cl(V,\mathfrak{q}) = 2^n$. Moreover, if $(e_i)_{i \in [n]}$ is an orthogonal basis of $V$, $(e_A)_{A \subseteq [n]}$ is an orthogonal basis of $\Cl(V,\mathfrak{q})$.
\end{theorem}

% Note that the Clifford algebra is a \textit{filtered algebra}.

% \begin{definition}
%     Define the \textit{multivector filtration} of $\Cl(V, \mathfrak{q})$ for grade $m \in \mathbb{N}_0$ as 
%     \begin{equation}
%         \Cl^{(\leq m)} (V,\mathfrak{q}) \coloneqq \pi\left (\T^{(\leq m)} (V) \right), ~ ~ ~ ~ ~ ~ ~ ~ ~ \T^{\leq m} (V) \coloneqq \bigoplus^m_{l = 0} V^{\otimes l}.
%     \end{equation}
%     One has a filtration on the space $\Cl(V,\mathfrak{q})$
%     \begin{equation}
%         \mathbb{F} = \Cl^{(\leq 0)} (V,\mathfrak{q}) \subseteq \Cl^{(\leq 1)} (V,\mathfrak{q}) \subseteq \Cl^{(\leq 2)} (V,\mathfrak{q}) \subseteq \ldots \subseteq \Cl^{(\leq n)} (V,\mathfrak{q}) = \Cl(V,\mathfrak{q}).
%     \end{equation}
% \end{definition}

% \begin{definition}[The grade of an element]
%     For $x \in \Cl(V,\mathfrak{q}) \setminus \{0\}$, we define its \textit{grade} by $k \in \{0,1,\ldots,n\}$ if 
%     \begin{equation}
%         x \in \Cl^{(\leq k)} (V,\mathfrak{q}) \setminus \Cl^{(\leq k-1)} (V,\mathfrak{q}).
%     \end{equation}
%     Denote $\operatorname{grd} x = k$ and define $\operatorname{grd} 0 \coloneqq - \infty$.
% \end{definition}
    
\begin{theorem}[The multivector grading of the Clifford algebra]
    Let $e_1, \ldots, e_n$ be an orthogonal basis of $(V, \mathfrak{q})$. Then for every $m = 0,\ldots, n$ we define the following vector subspace of $\Cl(V, \mathfrak{q})$:
    \begin{equation}
        \Cl^{(m)}(V,\mathfrak{q}) \coloneqq \operatorname{span} \{e_A ~ | ~ A \subseteq [n], |A| = m \}.
    \end{equation}
    Then the vector subspaces $\Cl^{(m)} (V,\mathfrak{q}), m = 0, \ldots, n$, are independent of the choice of the orthogonal basis. One has $\dim_{\mathbb{F}}(\Cl^{(m)}(V,\mathfrak{q})) = \binom{n}{m}$. Also, $\Cl^{(m)}(V,\mathfrak{q})$ is the space of all elements of degree $m$.
\end{theorem}

The direct sum
\begin{equation}
    \Cl(V,\mathfrak{q}) = \bigoplus^n_{m=0} \Cl^{(m)}(V,\mathfrak{q}),
\end{equation}
is an orthogonal sum. We write $x \in \Cl(V,\mathfrak{q})$ as $x = x^{(0)} + x^{(1)} + \ldots + x^{(n)}$, where $x^{(m)} \in \Cl^{(m)}(V,\mathfrak{q})$ denotes the grade-$m$ part of $x$. Define 
\begin{equation}
    \Cl^{[0]}(V,\mathfrak{q}) \coloneqq \bigoplus^n_{m \text { even}} \Cl^{(m)}(V,\mathfrak{q}), ~ ~ ~ ~ ~ ~ ~ ~ ~ \Cl^{[1]}(V,\mathfrak{q}) \coloneqq \bigoplus^n_{m \text { odd}} \Cl^{(m)}(V,\mathfrak{q}),
\end{equation}
where elements of even and odd parity, respectively. 
We also have an orthogonal decomposition of $\Cl(V,\mathfrak{q}$ which is the parity decomposition:
\begin{equation}
    \Cl(V,\mathfrak{q}) =  \Cl^{[0]}(V,\mathfrak{q})  \bigoplus \Cl^{[1]}(V,\mathfrak{q}).
\end{equation}
We write $x \in \Cl(V,\mathfrak{q})$ as $x = x^{[0]} + x^{[1]}$, where $x^{[i]} \in \Cl^{[0]}(V,\mathfrak{q})$, to denote the \textit{parity decomposition} of $x$.

\subsection{Clifford Group}
Let $\Cl^{\times}(V,\mathfrak{q})$ denote the group of invertible elements of the Clifford algebra. For $w \in \Cl^{\times}(V,\mathfrak{q})$, we defined the \textit{(adjusted) twisted conjugation}:
\begin{align}
    \rho(w) \colon \Cl(V,\mathfrak{q}) & \longrightarrow \Cl(V,\mathfrak{q}) \notag \\
    x \quad & \longmapsto wx^{[0]}w^{-1} + \alpha(w)x^{[1]}w^{-1},
\end{align}
where $\alpha$ is the \textit{main involution} of $\Cl^{\times}(V,\mathfrak{q})$, which is given by $\alpha(w) \coloneqq w^{[0]} - w^{[1]}$. 
The \textit{Clifford group}, denoted by $\Gamma(V,\mathfrak{q})$, is a subgroup of $\Cl^{\times}(V,\mathfrak{q})$ consists of elements that is parity homogeneous and preserves $V$ via $\rho$:
\begin{equation}
    \Gamma(V,\mathfrak{q}) = \Bigl\{ w \in \Cl^{\times}(V,\mathfrak{q}) \cap \left ( \Cl^{[0]}(V,\mathfrak{q}) \cup  \Cl^{[1]}(V,\mathfrak{q}) \right) ~ | ~  \rho(V) \subset V \Bigr\}.
\end{equation}

\begin{proposition}
    \begin{enumerate}
        \item For $w \in \mathbb{F}^{\times}$, we have $w \in \Gamma(V,\mathfrak{q})$ and $\rho(w) = \id_{\Cl(V,\mathfrak{q})}$.
        \item For $w \in V$ with $\mathfrak{q}(w) \neq 0$, we have $w \in \Gamma(V,\mathfrak{q})$ and $\rho(w)|_V$ is the reflection onto the hyperplane that is normal to w. $\rho(w)$ is given by the following formula: For $v \in V$:
        \begin{equation}
            \rho(w)(v) = v - 2\frac{\mathfrak{b}(w,v)}{\mathfrak{b}(w,w)}w.
        \end{equation}
    \end{enumerate}
\end{proposition}

\begin{theorem}
    The map 
    \begin{align}
        \rho \colon \Gamma(V,\mathfrak{q})  & \longrightarrow \Aut_{\mathbf{Alg},grd}(\Cl(V,\mathfrak{q})) \notag \\ 
        w \quad & \longmapsto \rho(w),
    \end{align}

    is a well-defined group homomorphism from the Clifford group to the group of $\mathbb{F}$-algebra automorphisms of $\Cl(V, \mathfrak{q})$ that preserve the multivector grading of $\Cl(V, \mathfrak{q})$. In particular, $\Cl(V,\mathfrak{q})$ and $\Cl^{(m)}(V,\mathfrak{q})$ for $m = 0,1,\ldots,n$ are group representations of $\Gamma(V,\mathfrak{q})$ via $\rho$.

    Moreover, each $\rho(w)$ generates an orthogonal map with respect to the (extended) bilinear form $\mathfrak{b}$.
\end{theorem}

\begin{theorem}[All grade projections are Clifford group equivariant]
\label{theorem:equivariantgradeprojection}
    For $w \in \Gamma(V,\mathfrak{q}), x \in \Cl(V,\mathfrak{q})$ and $m = 0, 1, \ldots, n$ we have the following equivariance property:
    \begin{equation}
        \rho(w)(x^{(m)}) = (\rho(w)(x))^{(m)}.
    \end{equation}
\end{theorem}

\begin{theorem}[All polynomials are Clifford group equivariant]
\label{theorem:equivariantpolynomial}
    Let $F \in \mathbb{F}[T_1, \ldots, T_l]$ be a polynomial in $l$ variables with coefficients in $\mathbb{F}$. Consider $w \in \Gamma(V,\mathfrak{q})$ and $l$ elements $x_1,\ldots,x_l \in \Cl(V,\mathfrak{q})$. We have the following equivariance property:
    \begin{equation}
        \rho(w)(F(x_1,\ldots,x_l)) = F(\rho(w)(x_1),\ldots,\rho(w)(x_l)).
    \end{equation}
\end{theorem}

\subsection{Orthogonal Group and its Action on Clifford Algebra}

\begin{definition}[Orthogonal group]
    Define the \textit{orthogonal group} of $(V,\mathfrak{q})$ as
    \begin{equation}
         \Ortho(V,\mathfrak{q}) = \Bigl \{  \text{linear automorphism } f \text{ of } V ~ | ~ \forall v \in V, \mathfrak{q}(f(v)) = \mathfrak{q}(v) \Bigr \}.
    \end{equation}
\end{definition}

\begin{theorem}[Theorem of Cartan-Dieudonné]
    Let $(V,\mathfrak{q})$ be a non-degenerate quadratic space of finite dimension $\dim V = n$ over a field $\mathbb{F}$ of $\operatorname{char} \mathbb{F} \neq 2$. Then every element $g \in \OO(V,\mathfrak{q})$ can be written as:
    \begin{equation}
        g = r_1 \circ \cdots \circ r_k,
    \end{equation}
    where $0 \le k \le n$ and $r_i$'s are reflections with respect to non-singular hyperplanes.
\end{theorem}

\begin{theorem}
    If $(V,\mathfrak{q})$ is non-degenerate then we have a short exact sequence:
    \begin{equation}
        \begin{tikzcd}
            0 \arrow{r} & \mathbb{F}^{\times} \arrow[hookrightarrow]{r}{i} & \Gamma(V,\mathfrak{q}) \arrow{r}{\rho|_{V}}  & \OO(V,\mathfrak{q}) \arrow{r}   & 0.
    \end{tikzcd}
    \end{equation}
    In particular, one has $\Gamma(V,\mathfrak{q}) / \mathbb{F}^{\times} \simeq \OO(V,\mathfrak{q})$.
\end{theorem}

The above implies that $\OO(V,\mathfrak{q})$ acts on whole $\Cl(V, \mathfrak{q})$ in a well-defined way. 
Concretely, if $x \in \Cl(V,\mathfrak{q})$ is of the form $x = \sum_{i \in I} c_i \cdot v_{i,1} \cdots v_{i,k_i}$ with $v_{i,j} \in V, c_i \in \mathbb{F}$ and $f \in \OO(V,\mathfrak{q})$ corresponds to $[w] \in \Gamma(V,\mathfrak{q}) / \mathbb{F}^{\times}$, then:
\begin{equation}
    \rho(w)(x) = \sum_{i \in I} c_i \cdot \rho(w)(v_{i,1}) \cdots \rho(w)(v_{i,k_i}) = \sum_{i \in I} c_i \cdot f(v_{i,1}) \cdots f(v_{i,k_i}).
\end{equation}

\begin{theorem}
    A map $f \colon \Cl(V,\mathfrak{q})^{p} \rightarrow \Cl(V,\mathfrak{q})^{q}$ is equivariant to the Clifford group $\Gamma(V,\mathfrak{q})$ if and only if it is equivariant to the orthogonal group $\OO(V,\mathfrak{q})$.
\end{theorem}

\subsection{Finite Dimensional Real Vector Space with Euclidean norm}
In this subsection, we only consider the case $\mathbb{F} = \mathbb{R}$ is the field of real numbers, and $V$ is an $n$-dimensional vector space over $\mathbb{R}$ with the quadratic form $\mathfrak{q}$ is the square of the Euclidean norm, i.e.  $\mathfrak{q}(\cdot) = \| \cdot \|_2^2$. 
In this case, $(V,\mathfrak{q})$ a quadratic space that is non-degenerate. The orthogonal group $\Ortho(V,\mathfrak{q})$ now is the usual orthogonal group $\Ortho(n)$.

\begin{proposition}
    $(V,\mathfrak{q})$ has an orthonormal basis.
\end{proposition}

\begin{proposition}
    For $w \in V, w \neq 0$, $\rho(w)$ defines the reflection of $V$ onto the hyperplane that is normal to $w$.
\end{proposition}

\begin{proposition}
    The Clifford group $\Gamma(V,\mathfrak{q})$ can be defined by:
    \begin{align}
        \Gamma(V,\mathfrak{q}) & = \{c \cdot v_1 \cdots v_k ~ | ~ c \in \mathbb{R}^{\times}, k \ge 0, v_i \in V, v_i \neq 0 \} \\ & = \{c \cdot v_1 \cdots v_k ~ | ~ c \in \mathbb{R}^{\times}, 0 \le  k \le n, v_i \in V, v_i \neq 0 \}.
    \end{align}
\end{proposition}

\section{Clifford Group Equivariant Layers}
\label{appendix:cliffordlayer}
This section provides a way to design a neural network where neurons are elements of Clifford Algebra $\Cl(V,\mathfrak{q})$ and equivariant under the action of the Clifford group $\Gamma(V,\mathfrak{q})$. Each layer has the form as follows:
\begin{equation}
    \mathbf{T}_{\Phi} \colon \Cl(V,\mathfrak{q})^p \rightarrow \Cl(V,\mathfrak{q})^q, \qquad x = (x_1, \ldots, x_p) \mapsto y = (y_1,\ldots,y_q),
\end{equation}
where $p,q$ represent the number of input and output channels, respectively, and $\Phi = (\phi_{-})$ is a learnable hyperparameter.

\begin{definition}[Linear Layer]
    The \textit{linear layer} $\mathbf{T}^{\operatorname{lin}}_{\Phi} \colon \Cl(V,\mathfrak{q})^p \rightarrow \Cl(V,\mathfrak{q})^q$ is defined as follows:
    \begin{equation}
        y_{c_{\operatorname{out}}}^{(k)} \coloneqq \sum^p_{c_{\operatorname{in}}=1} \phi_{c_{\operatorname{out}},c_{\operatorname{in}},k} \cdot x_{c_{\operatorname{in}}}^{(k)} ~ , ~ \forall c_{\operatorname{out}} = 1, \ldots, q.
    \end{equation}
\end{definition}

\begin{definition}[Geometric Product Layer]
    The interaction between two elements of Clifford Algebra 
 is defined as follows:
    \begin{equation}
        \mathbf{P}_{\Phi} \colon \Cl(V,\mathfrak{q}) \times \Cl(V,\mathfrak{q}) \rightarrow \Cl(V,\mathfrak{q}), \qquad \mathbf{P}_{\Phi}(x,\tilde{x})^{(k)} = \sum^n_{i=0} \sum^n_{j=0} \phi_{i,j,k} \cdot \left( x^{(i)} \tilde{x}^{(j)}\right)^{(k)}.
    \end{equation}

    We first apply a linear map to input $x = (x_1,\ldots,x_p)\in  \Cl(V,\mathfrak{q})^p$ to obtain $z = (z_1,\ldots,z_p) \in  \Cl(V,\mathfrak{q})^p$.
    
    The \textit{geometric product layer} $\mathbf{T}^{\operatorname{prod}}_{\Phi} \colon \Cl(V,\mathfrak{q})^p \rightarrow \Cl(V,\mathfrak{q})^p$ is defined as follows:
    \begin{equation}
        y^{(k)}_{c_{\operatorname{out}}} \coloneqq \mathbf{P}_{\phi_{c_{\operatorname{out}}}}(x_{c_{\operatorname{out}}},z_{c_{\operatorname{out}}})^{(k)} ~ , ~ \forall c_{\operatorname{out}} = 1, \ldots, p.
    \end{equation}
    
    The \textit{fully-connected geometric product layer} $\mathbf{T}^{\operatorname{prod}}_{\Phi} \colon \Cl(V,\mathfrak{q})^p \rightarrow \Cl(V,\mathfrak{q})^p$ is defined as follows:
    \begin{equation}
        y^{(k)}_{c_{\operatorname{out}}} \coloneqq \sum_{c_{in}=1}^{p} \mathbf{P}_{\phi_{c_{\operatorname{out}}, c_{\operatorname{in}}}}(x_{c_{\operatorname{in}}},z_{c_{\operatorname{in}}})^{(k)} ~ , ~ \forall c_{\operatorname{out}} = 1, \ldots, p.
    \end{equation}
\end{definition}

\begin{definition}[Normalization Layer]

The \textit{normalization layer} $\mathbf{T}^{\operatorname{norm}}_{\Phi} \colon \Cl(V,\mathfrak{q}) \rightarrow \Cl(V,\mathfrak{q})$ is defined as follows:
\begin{equation}
    x_{\text{out}}^{(m)} = \frac{x_{\text{in}}^{(m)}}{\sigma(\phi_m)\left(\mathfrak{q}\left(x_{\text{in}}^{(m)}\right)-1\right)+1}, ~ \forall m = 0, 1, \ldots,n.
\end{equation}
Here, $\sigma$ denotes the sigmoid function.
\end{definition}

\begin{definition}[Nonlinear Activation]
The \textit{nonlinear activation} $\mathbf{T}^{\operatorname{non-lin}} \colon \Cl(V,\mathfrak{q}) \rightarrow \Cl(V,\mathfrak{q})$ is defined as follows:
\begin{equation}
    x_{\text{out}}^{(0)} = \operatorname{ReLU}\left(x_{\text{in}}^{(0)}\right) \text{ and } x_{\text{out}}^{(m)} = \sigma \left(\mathfrak{q}\left(x_{\text{in}}^{(m)}\right) \right)x_{\text{in}}^{(m)}.
\end{equation}
Here, $\sigma$ denotes the sigmoid function. Note that, the nonlinear activation does not require learnable hyperparameter.
\end{definition}

The maps in the above definitions are equivariant to the Clifford group, by Theorem.~\ref{theorem:equivariantgradeprojection} and Theorem.~\ref{theorem:equivariantpolynomial}.

\section{Equivariance Proof for CG-EGNNs}
\label{appendix:equivariantproof}
We provide a proof for Theorem~\ref{theorem:equivariance} and also Theorem~\ref{theorem:enequivariant} that we recall below by:
\begin{equation}
	Q \cdot \mathbf{x}_i^{L}+g = \text{CG-EGNN}(Q \cdot \mathbf{x}_i+g),
\end{equation}
for all orthogonal matrix $Q \in \Ortho(n)$, translation vector $g \in \mathbb{R}^n$ and $i = 1, \ldots, M$.

For $\mathbf{x} = \{\mathbf{x}_1, \ldots, \mathbf{x}_M \}$, denote the mean-subtracted positions of $\mathbf{x}$ as follows:
\begin{equation}
    \overline{\mathbf{x}} = \left \{ \overline{\mathbf{x}}_1, \ldots,  \overline{\mathbf{x}}_M \right \},
\end{equation}
where
\begin{equation}
    \overline{\mathbf{x}}_i = \mathbf{x}_i - \frac{1}{M}\sum_{j=1}^M\mathbf{x}_j, \forall i = 1, \ldots, M.
\end{equation}
Now, we analyze CG-EGNN. Removing the residual connection and decomposing as follows:
\begin{align}
    \Psi \coloneq & ~ \text{CG-EGNN (without residual connection)} \\ = & ~  \mathbf{Projection} \circ \underbrace{\mathbf{Convolution} \circ \cdots \circ \mathbf{Convolution}}_{ L \text{ times}} \circ \mathbf{Embedding}.
\end{align}

Each map $\mathbf{Embedding}, ~ \mathbf{Convolution},~ \mathbf{Projection}$ is constructed by the layers presented in Appendix~\ref{appendix:cliffordlayer}, so they are $\Ortho(n)$-equivariant. So composition of them is $\Ortho(n)$-equivariant. Now we have:
\begin{align}
    \text{CG-EGNN}(Q \cdot \mathbf{x}_i+g) & = \left (Q \cdot \mathbf{x}_i+g \right ) +  \Psi \left (Q \cdot \mathbf{x}_i+g \right) \notag \\
    & \notag = (Q\cdot \mathbf{x}_i+g) +  \Psi(\overline{Q \cdot \mathbf{x}_i+g}) \\
    & \notag = (Q \cdot \mathbf{x}_i+g) + \Psi \left ( Q \cdot \mathbf{x}_i+g -  \frac{1}{M}\sum_{j=1}^M \left (Q \cdot \mathbf{x}_j +g  \right) \right ) \\
    & \notag = (Q \cdot \mathbf{x}_i+g) + \Psi \left ( Q \cdot \left(\mathbf{x}_i -  \frac{1}{M}\sum_{j=1}^M\mathbf{x}_j\right )\right) \\
    &  \notag = (Q \cdot \mathbf{x}_i+g) + \Psi \left ( Q \cdot \overline{\mathbf{x}}_i\right) \\
    &  \notag = (Q  \cdot \mathbf{x}_i+g) + Q\cdot\Psi \left ( \overline{\mathbf{x}}_i\right) \\
    & \notag = (Q\cdot \mathbf{x}_i+g) + Q\cdot\Psi \left ( \mathbf{x}_i\right) \\
    & \notag = Q\cdot \left (\mathbf{x} + \Psi \left ( \mathbf{x}_i\right) \right) + g \\
    & = Q\cdot \mathbf{x}_i^{L} + g.
\end{align}

The theorem is then proved.

\section{Proof of Theorem~\ref{thm:universal}}
\label{appendix:universalproof}

We provide a proof for Theorem~\ref{thm:universal}. Fix an arbitrary $\epsilon>0$. 
Since $\mathcal{X}$ is a compact metric space, $f$ is uniformly continuous on $\mathcal{X}$.
As a consequence, there exists $\delta>0$ such that: for arbitrary graphs $G$ and $G'$ in $\mathcal{X}$, we always have:
\begin{equation*}
d_H(G,G')<\delta \Rightarrow \left\Vert f(G)-f(G') \right\Vert_{\infty} < \epsilon.
\end{equation*}
Set $K > \max \left\{ \lceil \frac{1}{\delta} \rceil, \lceil \frac{1}{3\alpha} \rceil \right\}$.
Here, $\alpha$ is a positive number given in Eq.~\eqref{eq:X}.
Let $\mathcal{R} = \{\frac{2i-1}{2K} \,|\, i=1,\ldots,K\}$ be the set of equidistance values in the interval $[0,1]$.
Then $\mathcal{R}^d$ forms a lattice inside the $d$-dimensional box $[0,1]^d$.
For each point $z \in \mathbb{R}^{d}$, we define by $\bar{z}$ to be the nearest point of $z$ in the lattice $\mathcal{R}^{d}$.
For each $G \in \mathcal{X}$, we set $\bar{G} = \{\bar{z} \,|\, z \in G\}$ which is a finite set of points in the box $\mathcal{R}^d$.
According to the definition of $K$, different nodes of the same graph $G$ in $\mathcal{X}$ will have different images in $\mathcal{R}^d$.
Then we have $d_H(G,\bar{G}) < \delta$, 
thus, 
\begin{equation}\label{eq:barS}
\left\Vert f(G) - f(\bar{G}) \right\Vert_{\infty} < \epsilon.
\end{equation}

Next, we will find an alternative representation for $f(\bar{G})$.
First, for each $c \in \mathcal{R}^d$, we define a function $\delta_c \colon \mathbb{R}^d \to [0,1]$ by:
\begin{align}
	\delta_c(z) = \frac{1-e^{-d \left( z,\mathbb{R}^d \setminus B \left( c,\frac{1}{2K}\right) \right)}}{1-e^{-d \left( c,\mathbb{R}^d \setminus B \left( c,\frac{1}{2K}\right) \right)}}
\end{align}
for each $z \in \mathbb{R}^d$.
Then $\delta_c$ is a nonnegative continuous function on $\mathbb{R}^d$ such that:
\begin{equation}\label{eq:delta}
\delta_{c}(z) = \left\{
	\begin{aligned}
		&1, &&\text{if } z=c,\\
		& \in (0,1], &&\text{if } z \in B \big(c,\frac{1}{2K}\big),\\
		&0, &&\text{if } z \in \mathbb{R}^d \setminus B \big(c,\frac{1}{2K}\big).
	\end{aligned}
\right.
\end{equation}

Next, we define a map $\phi_m \colon \mathbb{R}^{d} \to [0,1]^{\mathcal{R}^d}$ as:
\begin{equation}
    \phi_m(z) = \left( \delta_{c}(z) \right)_{c \in \mathcal{R}^d}, \quad \text{for each } z \in \mathbb{R}^{d}.
\end{equation}
Then $\phi_m$ is a continuous function on $\mathbb{R}^d$ such that $\phi_m(\mathcal{R}^d) \subseteq \{0,1\}^{\mathcal{R}^d}$.
We also define a map $\tau \colon [0,1]^{\mathcal{R}^d} \to 2^{[0,1]^d}$ by:
\begin{equation}
\tau \left( (\epsilon_c)_{c \in \mathcal{R}^d} \right) = \{c \in \mathcal{R}^d \,|\, \epsilon_c > 0\},
\end{equation}
for each $(\epsilon_c)_{c \in \mathcal{R}^d} \in [0,1]^{\mathcal{R}^d}$.
Here, $2^{[0,1]^d}$ is the collection of all subsets of $[0,1]^d$.
We will need the following two claims:

\textbf{Claim 1.} For every graph $G \in \mathcal{X}$ and $c \in \mathcal{R}^d$, we have $\sum\limits_{z \in G} \delta_c(z) > 0$ if and only if $c \in \overline{G}$.

Indeed, since $\delta_c$ is a nonnegative function, we have 
$\sum\limits_{z \in G} \delta_c(z) > 0$ if and only if there exists $z_0 \in G$ such that $\delta_c(z_0)>0$.
Moreover, it follows from the definition of $\delta_c$ that 
$\delta_c(z_0)>0$ if and only if $z_0 \in B \big( c,\frac{1}{2K} \big)$. But $z_0 \in B \big( c,\frac{1}{2K} \big)$ if and only if $c = \overline{z_0} \in \overline{G}$.
The claim is proved.

\textbf{Claim 2.} For every graph $G \in \mathcal{X}$, we have:
\begin{equation}\label{eq:h-tau}
\tau \left(\sum_{z \in G} \phi_m(z)\right) = \bar{G}.
\end{equation}

Indeed, from the definition of $\phi_m$ and $\tau$, we have:
\begin{align}
\tau \left(\sum_{z \in G} \phi_m(z) \right) 
	&= \tau \left( \sum_{z \in G} (\delta_{c}(z))_{c \in \mathcal{R}^d} \right) \notag\\
	&= \tau \left( \left( \sum_{z \in G} \delta_{c}(z) \right)_{c \in \mathcal{R}^d} \right) \notag\\
&= \left\{ c \in \mathcal{R}^d \,|\, \sum_{z \in G} \delta_{c}(z)>0 \right\}.
\end{align}
Therefore, according to Claim 1, we have $\tau \left(\sum_{z \in G} \phi_m(z) \right) = \overline{G}$.
%According to the definition of $K$ and the function $\delta_c$ in Eq.~\eqref{eq:delta}, for a point $c \in \mathcal{R}^d$, the sum $\sum_{z \in G} \delta_{c}(z)$ is equal to $1$ if and only if there exists exactly one $z_0$ in $G$, which is the nearest point from $c$, such that $\delta_c(z_0)=1$.
%Therefore, it follows from the above equation that
%\begin{align*}
%\tau \left(\sum_{z \in G} \phi_e(z) \right) 
%	&= \left\{ c \in \mathcal{R}^d \,|\, \exists z \in G \text{ s.t. } c=\bar{z}\right\}\\
%	&= \bar{G}.
%\end{align*}
Claim 2 is then proved.

By using Claim 2 and Eq.~\eqref{eq:h-tau}, we obtain:
\begin{align}
f(\bar{G}) = f \circ \tau \left( \sum_{z \in G} \phi_m(z) \right).
\end{align}
Set $N=|\mathcal{R}^d|$ and $\phi_h = f \circ \tau$.
Here, in order to make the composition $f \circ \tau$ well-defined, we can restrict the domain of $\tau$ to a subset of $[0,1]^{\mathcal{R}^d}=[0,1]^N$ such that its image via $\tau$ is contained in $\mathcal{X}$.
We can extend $\phi_h$ to a continuous function on $\mathbb{R}^N$.
Then:
\begin{align}
f(\bar{G}) = \phi_h \left( \sum_{z \in G} \phi_m(z) \right).
\end{align}
It follows from Eq.~\eqref{eq:barS} that:
\begin{align}
\left\Vert f(G)- \phi_h \left( \sum_{z \in G} \phi_m(z) \right) \right\Vert_{\infty} < \epsilon.
\end{align}
The theorem is then proved.

\section{Implementation details}
\label{appendix:implementation_details}

In this session, we describe the experiment details for all experiments in session~\ref{sec:experiments}. All settings for CG-EGNN are kept as similar to the baseline as possible to ensure a fair comparison. All experiments in this paper is performed on a single A100 GPU.

\subsection{N-body System} 
\textbf{Experimental settings.} For this experiment, we employ the source code provided by~\citep{ruhe2023clifford}\footnote{\url{https://github.com/DavidRuhe/clifford-group-equivariant-neural-networks}} and follow the same settings in the paper. In which, the node features of the network consist of mean-subtracted positions of the particles, the charge of each particle and their initial velocities. The edge attributes are the
product of charges for connected node pairs. The goal of each model is to predict the displacement of each particle after $1000$ timesteps. To ensure translational invariance, we subtract the mean positions from the particles's positions. For all the baselines models, namely GNN~\citep{pmlr-v70-gilmer17a}, Tensor Field Network~\citep{thomas2018tensor}, SE(3)-Transformers~\citep{fuchs2020se}, Radial Field~\citep{kohler2020equivariant}, EGNN~\citep{satorras2021n}, SEGNN~\citep{brandstetter2021geometric}, and CGENN~\citep{ruhe2023clifford}, we use the results reported by~\citep{ruhe2023clifford}.

\textbf{Hyperparameter settings.} We maintain the set of hyperparamters similar to the set of hyperparamters specified in the source code of the original CGENN model. Especially, for CG-EGNN-$1$ and CG-EGNN-$1$-$2$, we set the learning rate to $\num{1e-3}$, weight decay to $\num{1e-4}$, hidden dimension to $20$, and the number of layer to $3$.  We train the model with Adam optimizer and cosine anealing learning rate scheduler~\citep{loshchilov2016sgdr}, batch size $100$ for $100000$ iterations, and we report the result on the test set at the point where the model checkpoint achieved the lowest MSE loss on the validation set.

\subsection{CMU Motion Capture with GMN settings.}

\textbf{Experimental settings.} In this experiment, we adapt the source code provided by~\citep{huang2022equivariant}\footnote{\url{https://github.com/hanjq17/GMN}}. Specifically, we consider the walking motion of the human subject (subject \#35) and adopt the random split used by~\citep{huang2022equivariant, han2022equivariant} with $495/498/498$ splits for training/validation/testing. The node features of the network contains initial velocities and positions of each joint. We augment the set of original edges representing the joints of the human figure with 2-hops neighbors following~\citep{huang2022equivariant, han2022equivariant}. The edge feature contains two components: the first component takes value $2$ if the edge is a two-hop edge and $1$ otherwise, the second component takes value $1$ when the edge is a stick and takes value $2$ when the edge is a hinge, and takes value $0$ otherwise. In this task, the prediction output corresponds to the location of each joint after 30 frames. For other baseline models, including: GNN~\citep{pmlr-v70-gilmer17a}, Tensor Field Network~\citep{thomas2018tensor}, SE(3)-Transformers~\citep{fuchs2020se}, Radial Field~\citep{kohler2020equivariant}, EGNN~\citep{satorras2021n}, GMN~\citep{huang2022equivariant}, we use the result reported from~\cite{huang2022equivariant}.

\textbf{Hyperparameter settings.} We keep the set of hyperparamters similar to the set of hyperparamters specified in the source code of the GMN~\citep{huang2022equivariant} paper. For CG-EGNN-$1$ and CG-EGNN-$1$-$2$, we found we can the hidden dimension can be reduced to $16$ without affecting the performance. We set the learning rate to $\num{5e-4}$, weight decay to $\num{1e-10}$, hidden dimension to $16$, and the number of layers to $4$. The model is trained with Adam optimizer for $500$ epochs with batch size $100$, and we report the performance of the model on the test set with the lowest validation loss.

% \subsection{CMU Motion Capture with EGHN~\cite{han2022equivariant} settings.}

\subsection{CMU Motion Capture with EGHN settings}

\textbf{Experimental settings.} We use the source code adapted from EGHN paper~\citep{han2022equivariant}\footnote{\url{https://github.com/hanjq17/EGHN}} for this experiment. Specifically, we use the same random split in~\citep{han2022equivariant}. The node features contains the velocities, positions and the augmented $z$-axis of each joint. The set of original edges are also augmented with 2-hops neighbors. The edge feature takes value $2$ if the edge is a two-hop edge and $1$ otherwise. In this task, the prediction output also corresponds to the location of each joint after 30 frames. Other baselines models in this experiment includes: GNN~\citep{pmlr-v70-gilmer17a}, Tensor Field Network~\citep{thomas2018tensor}, SE(3)-Transformers~\citep{fuchs2020se}, Radial Field~\citep{kohler2020equivariant}, EGNN~\citep{satorras2021n}, GMN~\citep{huang2022equivariant}, EGHN~\citep{han2022equivariant}. Asides from our model and CGENN, performance of other models follows from paper~\cite{han2022equivariant}.

\textbf{Hyperparameter settings.} In this experiment, we keep the set of hyperparamters similar to the set of hyperparamters specified in the source code of the GMN~\citep{han2022equivariant} paper. Specifically, for CG-EGNN-$1$ and CG-EGNN-$1$-$2$, we also decrease the hidden dimension to $8$, set the learning rate to $\num{5e-4}$, weight decay to $\num{1e-12}$, and fix the number of layers at $4$. The model is trained using the Adam optimizer, using early-stopping of 50 epochs.

\subsection{MD17 experiment.}
\label{appendix:implementation_details:MD17}
\begin{figure}
  \begin{center}
  % \vskip -0.1in
    \includegraphics[width=0.6\textwidth]{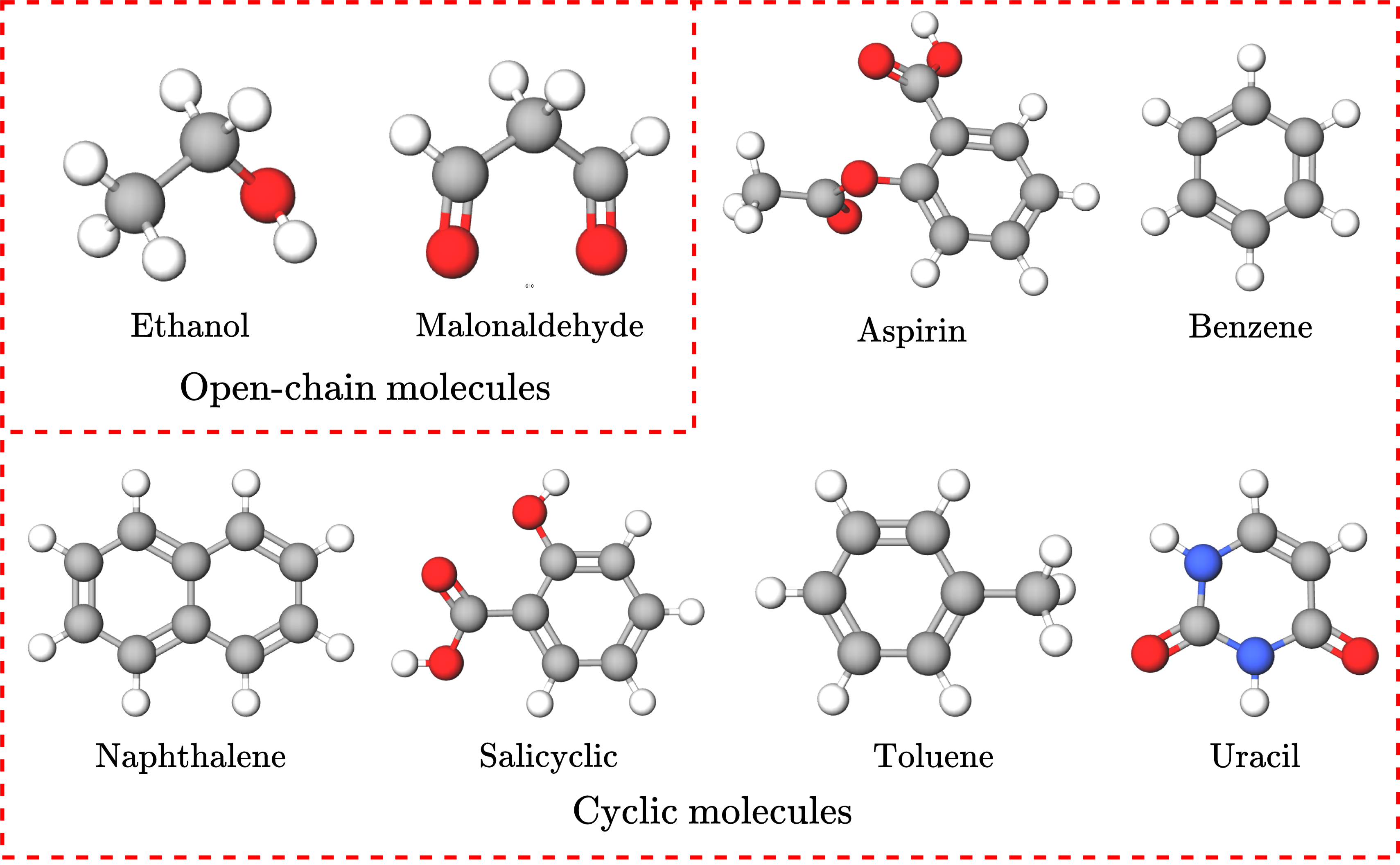}
  \end{center}
  % \vskip-0.25in
  \caption{Ball-and-stick model of molecules in MD17. Our method attains lower MSE for molecules with more complex structures.}
  \label{fig:molecules}
  % \vskip-0.3in
\end{figure}
\textbf{Experimental settings.}  In this experiment, we adopt the  MD17 dataset~\citep{chmiela2017machine}, which consists of trajectories of eight different molecules generated through molecular dynamics simulation. We utilize the source code adapted from~\citep{huang2022equivariant}\footnote{\url{https://github.com/hanjq17/GMN}} for this task and keep all settings the same. In which, we adopt the random split used by~\citep{huang2022equivariant} to divide the dataset into 50\% for training, 25\% for validation, and 25\% for testing. The time interval between the input and prediction frames is $T=5000$ timesteps. For this task, the feature node of each atom consists of the initial position, velocity and charges of the atom. The edge attribute is the concatenation of the atom number and its edge type indicator. The task of this experiment is to predict the future position of each atom given the current state of the molecule. We only consider the prediction for the position of large atoms and mask out all hydrogen atoms. Similar to~\citep{huang2022equivariant}, the graph of the molecule are also augmented with 2-hop neighbors. All the results of baseline models in this experiment are as reported from paper~\citep{huang2022equivariant}. 

\textbf{Hyperparameter settings.} In this experiment, we use the same hyperparameter setting for both CG-EGNN-$1$ and CG-EGNN-$1$-$2$ for all molecule tasks. In which, we set the learning rate to $\num{5e-4}$, weight decay to $\num{1e-10}$, reduce the hidden dimension to $8$, and set the number of layers to $4$. The models are trained with Adam optimizer for $500$ epochs with batch size $100$ and report the result on the test set at the checkpoint with the lowest MSE loss on the validation set.

\subsection{5D Convex Hulls}
\textbf{Experimental settings.} In this experiment, we utilize the source code provided by~\citep{ruhe2023clifford} and~\citep{liu2024clifford}\footnote{\url{https://github.com/congliuUvA/Clifford-Group-Equivariant-Simplicial-Message-Passing-Networks}} for the Convex Hulls experiment. The implementation of CGENN is adapted from~\citep{ruhe2023clifford}, the implementation of CSMPN~\cite{liu2024clifford} is adapted from~\citep{liu2024clifford}. Specifically, we generate a dataset of 16384 convex hulls with 8 vertices samples for each of the train, validation, and test set. Each convex hull is constructed by randomly sampling eight points from a standard normal distribution. To ensure a fair comparison, we keep the number of parameters in our implementations similar to~\cite{liu2024clifford} and maintain the same experimental settings. We compare our models with the following baselines: GNN~\citep{pmlr-v70-gilmer17a}, EGNN~\citep{satorras2021n}, CGENN~\citep{ruhe2023clifford}, EMPSN~\cite{eijkelboom2023n}, and CSMPN~\cite{liu2024clifford}.

\textbf{Hyperparameter settings.} In this experiment, we keep the set of hyperparamters similar to the set of hyperparamters specified in the source code of~\citep{liu2024clifford} paper. Specifically, for CG-EGNN-$1$ and CG-EGNN-$1$-$2$, we also decrease the hidden dimension to $16$, set the learning rate to $\num{1e-3}$, and fix the number of layers at $4$. The model is trained using the Adam optimizer, using early-stopping of $\num{1e5}$ iterations.

\subsection{Ablation study on the effect of including high order messages on 3D convex hull dataset}
\label{appendix:ablation_study}

% ----
\begin{table}[t]
\caption{MSE ($\times 10^{-2}$) of the 3D Convex Hulls experiment. The best CG-EGNN message orders combinations with $1, 2, 3$ elements are highlighted. The best model for all combination of message orders is CG-EGNN-$1$-$2$-$3$. }
\medskip\centering
\begin{adjustbox}{width=1.0\textwidth}
\small
\begin{tabular}{c c c c ccccccc }
\toprule
\multirow{2}{*}{\makecell[c]{Nodes\\in graphs}}& \multirow{2}{*}{GNN} & \multirow{2}{*}{EGNN} & \multirow{2}{*}{CGENN} & \multicolumn{7}{c}{CG-EGNN message orders} \\
\cmidrule{5-11}
&&&&$1$&$2$&$3$&$1-2$&$1-3$&$2-3$&$1-2-3$ \\
\midrule
$6$                                                                          & $6.472$                 & $7.347$                 & $6.251$                  &  {$2.531$} &  {$\mathbf{0.8410}$} &  {$1.057$} &  {$\mathbf{0.5604}$}   &  {$0.5683$}   &  {$0.7838$}   & $\mathbf{0.3970}$     \\
$7$                                                                          & $12.60$                     & $10.85$                     & $25.48$                      &  {$4.690$} &  {$\mathbf{2.379}$} &  {$2.865$} &  {$\mathbf{\mathbf{1.665}}$}   &  {$2.080$}   &  {$1.809$}   & $\mathbf{1.064}$     \\
$8$                                                                          & $24.45$                     & $12.05$                     & $55.47$                      &  {9.487} &  {$\mathbf{4.782}$} &  {$6.934$} &  {$\mathbf{4.042}$}   &  {$4.731$}   &  {$5.899$}   & $\mathbf{3.660}$    \\ \bottomrule
\end{tabular}
\end{adjustbox}
\label{tab:ablation_study}%
\end{table}

\begin{table}[t]
\caption{Runtime (second/it) of the 3D Convex Hulls experiment.}
\medskip\centering
\begin{adjustbox}{width=1.0\textwidth}
\begin{tabular}{c c c c ccccccc }
\toprule
\multirow{2}{*}{\makecell[c]{Nodes\\in graphs}}& \multirow{2}{*}{GNN} & \multirow{2}{*}{EGNN} & \multirow{2}{*}{CGENN} & \multicolumn{7}{c}{CG-EGNN message orders} \\
\cmidrule{5-11}
&&&&$1$&$2$&$3$&$1-2$&$1-3$&$2-3$&$1-2-3$ \\
\midrule
$6$                                                                          & $0.005421$                 & $0.005735$                 & $0.01792$                  &  {$0.1159$} &  {$0.1256$} &  {$0.1270$} &  {$0.1817$}   &  {$0.1838$}   &  {$0.1948$}   & $0.2554$     \\
$7$                                                                          & $0.005512$                     & $0.005831$                     & $0.01826$                      &  {$0.1170$} &  {$0.1496$} &  {$0.1739$} &  {$0.2140$}   &  {$0.2394$}   &  {$0.2722$}   & $0.3365$     \\
$8$                                                                          & $0.005785$                     & $0.006169$                     & $0.01923$                      &  {0.1234} &  {$0.1962$} &  {$0.2944$} &  {$0.2669$}   &  {$0.3664$}   &  {$0.4374$}   & $0.5124$    \\ \bottomrule
\end{tabular}
\end{adjustbox}
\label{tab:ablation_study_runtime}%
\end{table}

% ----
\textbf{Experimental settings.} 
 For this experiment, we create 3 dataset of 3D convex hulls with number of nodes per graph $\in \{6, 7, 8\}$. Each vertice is sampled from a standard normal distribution to obtain 3 dataset of $4000/4000/4000$ samples for training/validation/test each. We run all 3 dataset for baselines GNN~\citep{pmlr-v70-gilmer17a}, EGNN~\citep{satorras2021n}, CGENN~\citep{ruhe2023clifford}, and CG-EGNN models with all possible combination of message orders up to message order 3 (namely, the following list of combinations: \{$1$, $2$, $3$, $1-2$, $1-3$, $2-3$, $1-2-3$\}).

\textbf{Hyperparameter settings.} We set the hidden dimensions of GNN, EGNN, CGENN to $32$ and hidden dimensions of CG-EGNN models of all order combinations to $8$. The learning rate is set to $\num{1e-3}$, number of layers is $4$, all models are trained using Adam optimizer with early-stopping for $\num{5e4}$ iterations.

\section{Broader Impact}\label{appendix:broader-impact}
The introduction of Clifford Group Equivariant Graph Neural Networks (CG-EGNNs) in this paper holds significant societal impact by advancing the capabilities of neural networks to handle data symmetry, particularly in applications requiring precise geometric representations. This innovation has the potential to drive progress in fields such as drug discovery, materials science, and robotics, where accurate modeling of molecular and physical systems is essential. By enhancing the expressive power and maintaining equivariance properties, CG-EGNNs can lead to more accurate simulations and predictions, reducing the time and cost associated with experimental procedures. Furthermore, the improved performance on benchmarks like n-body simulations, CMU motion capture, and MD17 showcases the practical utility of CG-EGNNs in real-world scenarios. As these advanced neural networks become more integrated into scientific research and industry, they could facilitate breakthroughs in developing new materials, understanding complex biological processes, and creating sophisticated autonomous systems, ultimately contributing to technological advancements and improving quality of life.

\end{document}